\documentclass{article}

\usepackage[final]{neurips_2021}

\usepackage{amsmath, amssymb, amsthm}
\usepackage{mathtools}
\usepackage{bbm}
\usepackage{dsfont}

\usepackage[usenames,dvipsnames]{xcolor}
\definecolor{shadecolor}{gray}{0.9}

\usepackage{graphicx}
\usepackage[labelfont=bf]{caption}
\usepackage[format=hang]{subcaption}
\usepackage{stfloats}

\usepackage{booktabs, array}
\usepackage{multirow}
\usepackage{rotating}
\usepackage{enumitem}
\usepackage{xfrac}

\usepackage{soul}
\usepackage{algorithm,algorithmicx,algpseudocode}

\usepackage{listings}
\usepackage{fancyvrb}
\fvset{fontsize=\small}

\usepackage[colorlinks,linktoc=all]{hyperref}
\usepackage[all]{hypcap}
\hypersetup{citecolor=MidnightBlue}
\hypersetup{linkcolor=MidnightBlue}
\hypersetup{urlcolor=MidnightBlue}
\usepackage[nameinlink,capitalise]{cleveref}

\usepackage[acronym,smallcaps,nowarn,section,nogroupskip,nonumberlist]{glossaries}

\glsdisablehyper{}

\usepackage{wrapfig}

\usepackage{tikz}
\usetikzlibrary{tikzmark}

\lstdefinestyle{mystyle}{
    commentstyle=\color{OliveGreen},
    numberstyle=\tiny\color{black!60},
    stringstyle=\color{BrickRed},
    basicstyle=\ttfamily\scriptsize,
    breakatwhitespace=false,
    breaklines=true,
    captionpos=b,
    keepspaces=true,
    numbers=none,
    numbersep=5pt,
    showspaces=false,
    showstringspaces=false,
    showtabs=false,
    tabsize=2
}
\usepackage{xspace}

\makeatletter
\DeclareRobustCommand\onedot{\futurelet\@let@token\@onedot}
\def\@onedot{\ifx\@let@token.\else.\null\fi\xspace}

\def\ie{\emph{i.e}\onedot}

\makeatother

\newacronym{kd}{kd}{Knowledge Distillation}
\newacronym{be}{be}{BatchEnsemble}
\newacronym{de}{de}{Deep Ensemble}
\newacronym{bma}{bma}{Bayesian Model Average}
\newacronym{acc}{acc}{Accuracy}
\newacronym{nll}{nll}{Negative Log-Likelihood}
\newacronym{ece}{ece}{Expected Calibration Error}
\newacronym{bs}{bs}{Brier Score}
\newacronym{dee}{dee}{Deep Ensemble Equivalent}
\newacronym{ods}{ods}{Output Diversified Sampling}
\newacronym{roc}{roc}{Receiver Operating Characteristic}
\newacronym{snr}{snr}{Signal-to-Noise Ratio}
\newacronym{mimo}{mimo}{Multi-Input Multi-Output}
\newacronym{endd}{end$^2$}{Ensemble Distribution Distillation}

\newcommand{\calD}{{\mathcal{D}}}

\newcommand{\calF}{{\mathcal{F}}}

\newcommand{\calL}{{\mathcal{L}}}

\newcommand{\calN}{{\mathcal{N}}}

\newcommand{\calS}{{\mathcal{S}}}
\newcommand{\calT}{{\mathcal{T}}}

\newcommand{\bbE}{\mathbb{E}}

\newcommand{\bbR}{\mathbb{R}}

\newcommand{\bsr}{\boldsymbol{r}}
\newcommand{\bss}{\boldsymbol{s}}

\newcommand{\bsw}{\boldsymbol{w}}
\newcommand{\bsx}{\boldsymbol{x}}
\newcommand{\bsy}{\boldsymbol{y}}

\newcommand{\bsI}{\boldsymbol{I}}

\newcommand{\bsR}{\boldsymbol{R}}

\newcommand{\bsW}{\boldsymbol{W}}

\newcommand{\bvarepsilon}{{\boldsymbol{\varepsilon}}}

\newcommand{\btheta}{{\boldsymbol{\theta}}}

\theoremstyle{plain}%

\theoremstyle{definition}

\theoremstyle{remark}

\DeclareMathOperator*{\argmax}{arg\,max}
\DeclareMathOperator*{\argmin}{arg\,min}

\newcommand{\tr}{^\top}

\newcommand{\iidsim}{\overset{\mathrm{i.i.d.}}{\sim}}

\def\[#1\]{\begin{align}#1\end{align}}
\newcommand{\norm}[1]{\left\lVert#1\right\rVert}

\newcommand{\spm}[1]{\scriptstyle{\pm#1}}

\newcommand{\tbx}{\tilde{\bsx}}

\newcommand{\Lce}{\calL_{\text{CE}}}
\newcommand{\Lkd}{\calL_{\text{KD}}}
\newcommand{\Ljm}{\calL_{\text{JM}}}
\newcommand{\softmax}{\mathrm{softmax}}

\usepackage[utf8]{inputenc} 
\usepackage[T1]{fontenc}    
\usepackage{hyperref}       
\usepackage{url}            
\usepackage{booktabs}       
\usepackage{amsfonts}       
\usepackage{nicefrac}       
\usepackage{microtype}      
\usepackage{xcolor}         

\title{Diversity Matters When Learning From Ensembles}

\author{
    Giung~Nam$^{1 *}$,
    Jongmin~Yoon$^{1 *}$,
    Yoonho~Lee$^{2,3}$,
    Juho~Lee$^{1,2}$ \\
    KAIST$^1$, Daejeon, South Korea, AITRICS$^2$, Seoul, South Korea, Stanford University$^3$, USA \\
    \texttt{\{giung,\,jm.yoon,\,juholee\}@kaist.ac.kr} \\
}

\newcommand\blfootnote[1]{
    \begingroup
    \renewcommand\thefootnote{}\footnote{#1}
    \addtocounter{footnote}{-1}
    \endgroup
}

\begin{document}
\blfootnote{$^*$ Equal contribution}

\maketitle

\begin{abstract}
    Deep ensembles excel in large-scale image classification tasks both in terms of prediction accuracy and calibration.
    Despite being simple to train, the computation and memory cost of deep ensembles limits their practicability.
    While some recent works propose to distill an ensemble model into a single model to reduce such costs, there is still a performance gap between the ensemble and distilled models.
    We propose a simple approach for reducing this gap, i.e., making the distilled performance close to the full ensemble.
    Our key assumption is that a distilled model should absorb as much function diversity inside the ensemble as possible.
    We first empirically show that the typical distillation procedure does not effectively transfer such diversity, especially for complex models that achieve near-zero training error.
    To fix this, we propose a perturbation strategy for distillation that reveals diversity by seeking inputs for which ensemble member outputs disagree.
    We empirically show that a model distilled with such perturbed samples indeed exhibits enhanced diversity, leading to improved performance.
\end{abstract}


\section{Introduction}
\label{sec:introduction}

\gls{de}~\citep{lakshminarayanan2017simple}, a simple method to ensemble the same model trained multiple times with different random initializations, is considered to be a competitive method for various tasks involving deep neural networks both in terms of prediction accuracy and uncertainty estimation. 
Several works have tried to reveal the secret behind \gls{de}'s effectiveness. 
As stated by \citet{duvenaud2016early} and \citet{wilson2020bayesian}, \gls{de} can be considered as an approximate \gls{bma} procedure. 
\citet{fort2019de} studied the loss landscape of \glspl{de} and showed that the effectiveness comes from the diverse modes reached by ensemble members, making it well suited for approximating \gls{bma}. 
It is quite frustrating that most sophisticated approximate Bayesian inference algorithms, especially the ones based on variational approximations, are not as effective as \glspl{de} in terms of exploring various modes in parameter spaces.

Despite being simple to train, \gls{de} incurs significant overhead in inference time and memory usage. 
It is therefore natural to develop a way to reduce such costs. 
An example of such a method is \gls{kd}~\citep{hinton2015distilling}, which transfers knowledge from a large \textit{teacher network} to a relatively smaller \textit{student network}.
The student network is trained with a loss that encourages it to copy the outputs of the teacher networks evaluated at the training data. 
With \gls{kd}, there have been several works that learn a single student network by distilling from \gls{de} teacher networks. 
A na\"ive approach would be to directly distill ensembled outputs of \gls{de} teachers to the single student network. 
A better way proven to be more effective is to set up a student network having multiple subnetworks (multiple heads~\citep{tran2020hydra} or rank-one factors~\citep{mariet2020bekd}) and distill the outputs of each ensemble member to each subnetwork in a one-to-one fashion.
Nevertheless, the empirical performance of such distilled networks is still far inferior to \gls{de} teachers.

\begin{figure}[t]
    \centering
    \begin{subfigure}{\linewidth}
        \includegraphics[width=\linewidth]{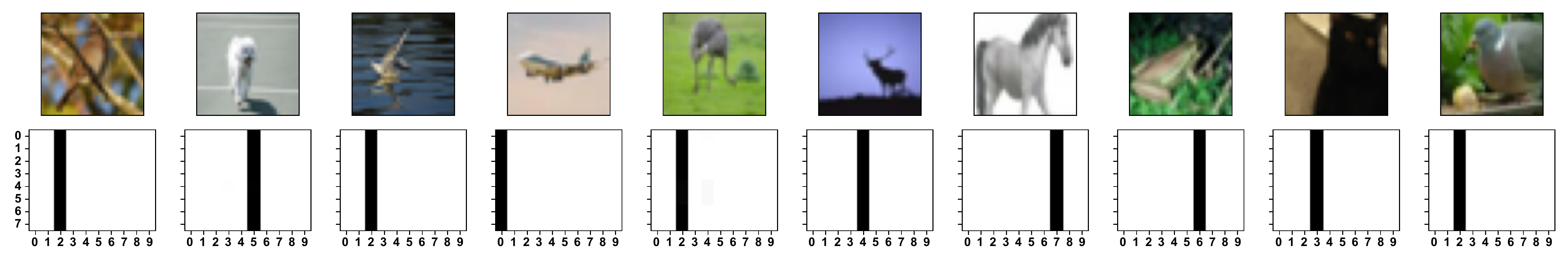}
    \end{subfigure}
    \caption{It shows randomly selected train examples from CIFAR-10 and their corresponding predictions given by 8 ResNet-32 models. In the second row, the vertical axis denotes the ensemble members, and the horizontal axis denotes the class index.}
    \label{fig:preview_predictions}
\end{figure}

There may be many reasons why the students are not doing as well as \gls{de} teachers.
We argue that a critical limitation of current distillation schemes is that they do not effectively transfer the diversity of \gls{de} teacher predictions to students. 
Consider an ensemble of deep neural networks trained for an image classification task. 
Given sufficient network capacity, each ensemble member is likely to achieve near-zero training error, meaning that the resulting outputs of ensemble members evaluated at the training set will be nearly identical, as shown in \cref{fig:preview_predictions}. 
In such a situation, reusing the training data during distillation will not encourage the student to produce diverse predictions.
This is quite critical, considering various results establishing that the effectiveness of \gls{de} comes from averaging out its diverse predictions.

To this end, we propose a method that amplifies the diversity of students learning from \gls{de} teachers. 
Our idea is simple: instead of using the same training set, use a perturbed training set for which the predictions of ensemble members are likely to be diversified. 
To implement this idea, we employ \gls{ods}~\citep{tashiro2020diversity}, a sampling scheme that finds small input perturbations that result in significant changes in outputs. Empirically, we confirm that the inputs perturbed with \gls{ods} result in large deviations in the \gls{de} teacher outputs. We further justify our method by analyzing the role of \gls{ods} perturbation. 
Specifically, we show that distilling using training data perturbed with \gls{ods} can be interpreted as an approximate Jacobian matching procedure where \gls{ods} improves the sample efficiency of the approximate Jacobian matching. 
Intuitively, by approximately matching Jacobians of teachers and students, we are transferring the outputs of teacher networks evaluated not only on the training data points but also on nearby points.

Using standard image classification benchmarks, we empirically validate that our distillation method promotes diversities in student network predictions, leading to improved performance, especially in terms of uncertainty estimation. 

\section{Backgrounds}
\label{sec:backgrounds}

\subsection{Settings and notations}
\label{subsec:settings_and_notations}

The focus of this paper is on the $K$-way classification problem taking $D$-dimensional inputs. We denote a student neural network as $\calS(\bsx):\bbR^D \to [0,1]^K$ and a teacher neural network as $\calT(\bsx): \bbR^D \to [0,1]^K$. $\calS(\bsx)$ and $\calT(\bsx)$ outputs class probabilities, and we denote the logits before softmax as $\hat\calS(\bsx)$ and $\hat\calT(\bsx)$. The $k$th element of an output is denoted as $\calS^{(k)}(\bsx)$. For \gls{de} teachers with $j=1,\dots, M$ members, the $j$th ensemble member is denoted as $\calT_j(\bsx)$. For a student network having $M$ subnetworks, the $j$th subnetwork is denoted as $\calS_j(\bsx)$.

\begin{figure}[t]
    \centering
    \includegraphics[width=1.0\linewidth]{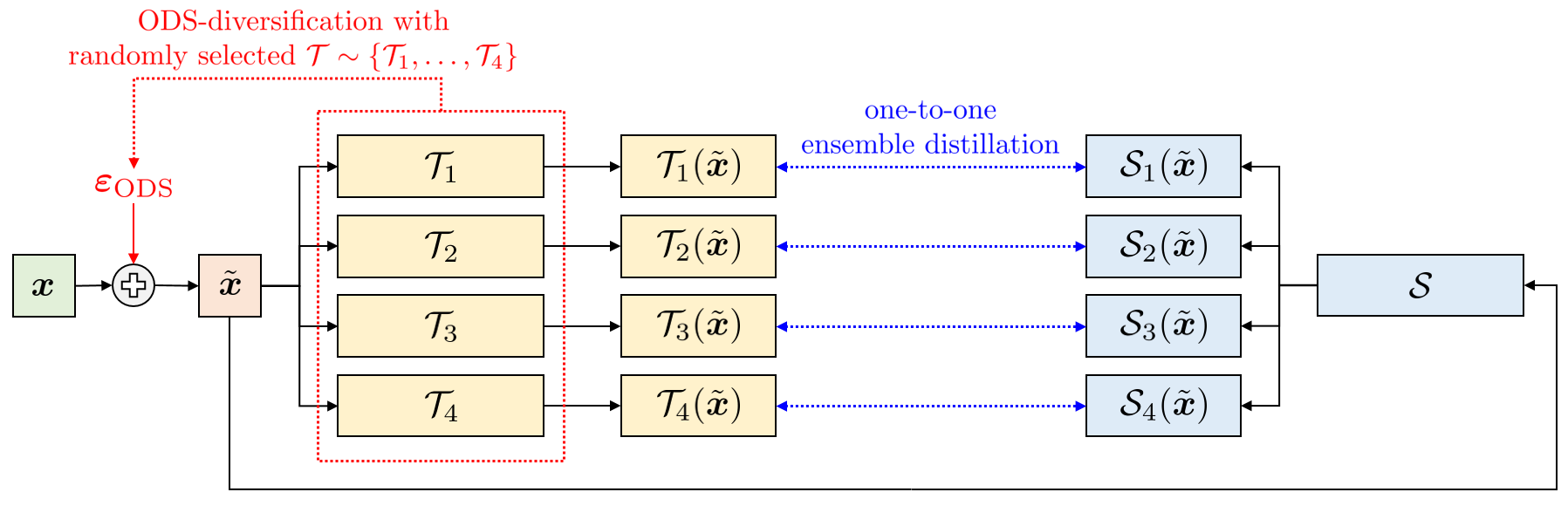}
    \caption{Overall structure of our method:
    (1) We first perturb train examples with respect to deep ensemble (\gls{de}) teachers, \ie, $\bsx \rightarrow \tilde{\bsx}$.
    (2) \gls{de} teachers produce \textit{diverse} predictions given perturbed train examples.
    (3) Since the diversity of teachers has been revealed, a student can effectively learn its diversity in the knowledge distillation framework.}
    \label{fig:overall_structure}
\end{figure}

\subsection{Knowledge distillation}
\label{subsec:knowledgedistillation}

Knowledge distillation (\gls{kd})~\citep{hinton2015distilling} aims to train a student network $\calS(\bsx)$ by matching its outputs to the outputs of a teacher network $\calT(\bsx)$. In addition to the standard cross-entropy loss between student outputs and the ground-truth labels, \ie,
$
    \Lce(\calS(\bsx), \bsy) =
    -\sum_{k=1}^K y^{(k)} \log \calS^{(k)}(\bsx),
$
\gls{kd} additionally encourages the student's output to be as close as possible to that of the teacher by also minimizing the KL-divergence between output class distributions of the student and the teacher:
\[
    \Lkd(\calS(\bsx), \calT(\bsx); \tau) =
    -\tau^2\sum_{k=1}^K \softmax^{(k)}\bigg(
        \frac{\hat\calT(\bsx)}{\tau}
    \bigg) \log \softmax^{(k)} \bigg(
        \frac{\hat\calS(\bsx)}{\tau}
    \bigg).
\]
Here, $\tau > 0$ is a \emph{temperature} parameter which controls the smoothness of the distributions. 
We scale the KL divergence by $\tau^2$ since the soft targets scale as $1 / \tau^2$.
As a result, in the \gls{kd} framework, the student imitates the teacher's outputs for training examples by minimizing $\calL = (1-\alpha) \Lce + \alpha \Lkd$, where $\alpha \in (0,1)$ is a hyperparameter to be specified.

\subsection{BatchEnsemble and one-to-one distillation}
\label{subsec:batchensemble}

\gls{be}~\citep{wen2019batchensemble} is a lightweight ensemble method that reduces the number of parameters by weight sharing.
Specifically, each layer of a \gls{be} model consists of shared weights $\bsW$ and rank-one factors $\bsr_j\bss_j\tr$ for $j=1,\dots, M$ with $M$ being number of subnetworks.
Unlike the full ensemble that constructs full set of $M$ weights, \gls{be} shares $\bsW$ across all subnetworks while the rank-one factors are kept specific to each subnetwork.
The weight matrix of the $j$th subnetwork is computed as $\bsW \circ \bsr_j\bss_j\tr$ and consequently \gls{be} requires far fewer parameters than a full ensemble.

During training, \gls{be} simultaneously trains all subnetworks by feeding a duplicate of the current minibatch for each rank-one factor $\bsr_j\bss_j\tr$ into the network.
This procedure is easily vectorized because the computation graph for each subnetwork has the same structure, differing only in weight values.
\citet{mariet2020bekd} proposed to further improve the performance of \gls{be} by distilling from fully trained \gls{de} teachers instead of training from scratch.
During distillation, the parameters of $j$th subnetwork $\calS_j(\bsx)$ is trained by distilling from the $j$th \gls{de} teacher $\calT_j(\bsx)$. The corresponding objective function is then written as (c.f., \cref{fig:overall_structure} without perturbation)
\[
    \calL = \sum_{j=1}^M \Big\{
        (1-\alpha)\Lce(\calS_j(\bsx),\bsy) + \alpha\Lkd(\calS_j(\bsx),\calT_j(\bsx);\tau)
    \Big\}.
\]
Throughout the rest of the paper, we will use the \gls{be} as student networks and follow this one-to-one distillation framework.

\subsection{\texorpdfstring{Output diversified sampling (\gls{ods})}{Output diversified sampling (ODS)}}
\label{subsec:ods}

\Gls{ods}~\citep{tashiro2020diversity} is a sampling method that can maximize diversities in outputs of a given function; it is originally introduced to replace random input perturbation to enhance the performance of the adversarial attack.
Let $\calF : \bbR^D \to \bbR^K$ be a target function.
\gls{ods} draws a perturbation vector guided from the \emph{gradient} of the target function as follows.
\[\label{eq:ods_def}
    \bsw \sim \mathrm{Unif}([-1,1])^K, \quad
    \bvarepsilon_{\text{ODS}}(\bsx, \calF, \bsw) = \frac{
        \nabla_{\bsx}(\bsw\tr\calF(\bsx))
    }{
        \norm{\nabla_{\bsx}(\bsw\tr\calF(\bsx))}_2
    } \in \bbR^D.
\]
The intuition behind this sampling strategy is to seek the direction in the input space that maximizes the similarity between the function output and a randomly sampled vector $\bsw$.
Following such a direction can make diverse output shifts due to the randomness in $\bsw$.


\section{Learning from Ensembles with Output Diversification}
\label{sec:methods}

\subsection{\texorpdfstring{One-to-one distillation with \gls{ods}}{One-to-one distillation with ODS}}

As discussed before, vanilla one-to-one distillation evaluates \gls{de} teachers on points where their outputs are nearly identical, so the students trained using them tend to exhibit low diversity in outputs.
To avoid this undesirable behavior, we propose to instead train students on points perturbed with \gls{ods}.
Let $(\bsx, \bsy)$ be an training input pair.
We first pick a \emph{random teacher} $\calT_r$ uniformly from $\{\calT_1,\dots, \calT_M\}$ and perturb $\bsx$ as
\[
    \tbx = \bsx + \eta \bvarepsilon_{\text{ODS}}(\bsx, \softmax(\hat\calT_r(\bsx)/\tau), \bsw),
\]
where $\eta > 0$ is a step-size.
Ideally, we would like to evaluate \gls{ods} vectors for each teacher to build $M$ perturbed versions of $\bsx$, but this would take too much time for gradient computation, especially when we have a large number of ensemble models.
Instead, we just pick one of the teachers, use it to perturb the input, and share the perturbed input to train all students with the following loss function,
\[\label{eq:kl_ods_loss}
    \calL = \sum_{j=1}^M \Big\{
        (1-\alpha)\Lce(\calS_j(\bsx),\bsy) + \alpha\Lkd(\calS_j(\tbx),\calT_j(\tbx);\tau)
    \Big\},
\]
where we are evaluating the KL-divergence between teacher outputs and student outputs on the \emph{perturbed input}.

An implicit assumption here is that \emph{a diversifying direction computed from a specific random teacher can bring diversities in the outputs of all the teacher networks.}
This is based on the \emph{transferability} assumption empirically justified in \citet{tashiro2020diversity}: \gls{ods} perturbations are computed on a surrogate model when we do not have access to the true model, assuming that the Jacobians computed from the surrogate model can approximate the true Jacobian to some extent.
In our case, let $\calT_1$ be a randomly picked teacher with which an \gls{ods} perturbation is computed, and $\calT_2$ be another teacher. 
Assuming the transferability of Jacobian, we can let $\nabla_{\bsx} \calT_1(\bsx) \approx \bsR \nabla_{\bsx}\calT_2(\bsx)$ for some matrix $\bsR$.
Then the \gls{ods} perturbation computed from $\calT_1(\bsx)$ is
\[
\bvarepsilon_\text{ODS}(\bsx, \calT_1, \bsw) \propto \bsw\tr \nabla_{\bsx} \calT_1(\bsx) 
\approx (\bsR\tr\bsw)\tr \nabla_{\bsx} \calT_2(\bsx) \propto \bvarepsilon_\text{ODS}(\bsx, \calT_2, \bsR\tr\bsw).
\]
That is, the \gls{ods} perturbation computed from $\calT_1$ acts as another \gls{ods} perturbation with different random guide vector $\bsR\tr\bsw$.
Hence, the same perturbation drives the outputs of $\calT_1$ and $\calT_2$ towards different directions.
Note that without transferability, the Jacobians $\nabla_{\bsx} \calT_1(\bsx)$ and $\nabla_{\bsx} \calT_2(\bsx)$ would be vastly different and  so are the directions $\bsw$ and $\bsR\tr\bsw$.
This may be good in terms of diversification, but the resulting perturbed input act as an adversarial example not very useful for distillation, making the teachers completely disagree (refer to~\cref{subsec:app_adversarial} for more details).
Our experiments in~\cref{subsec:exp_ods_diversity} and \cref{subsec:exp_ods_grad} empirically re-confirm this transferability assumption.

We can also interpret the random teacher selection as a stochastic approximation.
Specifically, to find a direction maximizing diversities between teachers, we can consider an objective
\[
\bbE_{\bsw_1,\dots, \bsw_M}\bigg[\sum_{j=1}^M \bsw_j\tr \calT_j(\bsx) \bigg], 
\]
with $\bsw_1, \dots, \bsw_M \iidsim \mathrm{Unif}([-1,1])^K$.
That is, we want to find a direction in the input space making the teachers $\{\calT_j\}_{j=1}^M$ follow different guide vectors $\{\bsw_j\}_{j=1}^M$. 
Picking a random teacher and computing the \gls{ods} perturbation can be understood as a stochastic approximation of the gradient of this objective.
Again, thanks to the transferability of the Jacobians, the variance of this stochastic gradient would be small, making it sufficient to use a single random teacher.

In addition, we empirically found that scaling the \gls{ods} updates according to confidence slightly improves performance.
That is, we perturb an input $\bsx$ as
\[\label{eq:confods}
    \tbx = \bsx + \eta C_{\text{max}}(\bsx, \calT_r, \tau) \bvarepsilon_{\text{ODS}}(\bsx, \softmax(\hat\calT_r(\bsx)/\tau), \bsw),
\]
where $C_\text{max}(\bsx, \calT_r, \tau) = \max_k \softmax^{(k)}(\hat\calT_r(\bsx)/\tau)$ denotes the maximum class probability.
This scheme takes larger steps towards \gls{ods} directions for datapoints with high confidence.
We call this modification Conf\gls{ods}.

The training procedure with \gls{ods} perturbation is straightforward (\cref{alg:method}); we can just add the \gls{ods} computation procedure and replace the loss function from the original \gls{kd} training procedure.
\cref{fig:overall_structure} depicts our overall training procedure as a diagram.

\subsection{Interpretation as approximate Jacobian matching}
\label{subsec:approx_grad}

We can also interpret our distillation strategy as an approximate Jacobian matching procedure. \citet{srinivas2018jacobian} shows that a \gls{kd} procedure on inputs perturbed by small noise implicitly encourages matching the Jacobians of a teacher and a student.
More specifically, let $\calS$ and $\calT$ be student and teacher models.
The first-order Taylor expansion of the expected \gls{kd} loss on perturbed inputs can be written as%
\footnote{We assume $\tau=1$ for notational simplicity, but the argument applies equally well to any $\tau$.}
\[
    \bbE_{\bvarepsilon}[
        \Lkd(\calS(\bsx+\bvarepsilon),\calT(\bsx+\bvarepsilon)
    ] = \Lkd(\calS(\bsx),\calT(\bsx)) + \Ljm(\calS(\bsx),\calT(\bsx)) + o(\sigma^2),
\]
where $\sigma^2 = \bbE[\bvarepsilon\tr\bvarepsilon]$ and 
\[\label{eq:jm_loss}
    \Ljm(\calS(\bsx),\calT(\bsx)) = -\bbE_{\bvarepsilon}\left[
        \sum_{k=1}^{K} \frac{
            \bvarepsilon\tr
            \nabla_{\bsx}\calT^{(k)}(\bsx)
            \nabla_{\bsx}\tr \calS^{(k)}(\bsx)
            \bvarepsilon 
        }{
            \mathcal{S}^{(k)}(x)
        } 
    \right].
\]
While \citet{srinivas2018jacobian} suggested using isotropic Gaussian noise $\bvarepsilon \sim \calN(\boldsymbol{0}, \sigma^2\bsI_D)$, we argue that simple Gaussian noise is ineffective, especially for high-dimensional data that are likely to be embedded in low-dimensional manifolds.
In such a case, most perturbation directions are orthogonal to the data manifold, and thus Gaussian noise becomes uninformative.
On the other hand, \gls{ods} directly uses the Jacobian of the teacher to construct the perturbation direction.
We conjecture that due to this fact, \gls{kd} with \gls{ods} perturbation is a much more sample-efficient approximate Jacobian matching procedure compared to using Gaussian perturbations.
We perform experiments to empirically validate this claim in \cref{subsec:exp_ods_grad}.

The above analysis is for the usual \gls{kd} framework with a single teacher and a single student.
In our case, we are matching $M$ teachers to $M$ student subnetworks, but with \gls{ods} perturbations computed from a single teacher randomly sampled from $M$ teachers.
Again, we are operating under the assumption that the Jacobian computed from a specific teacher will transfer well to other teachers, so the approximate Jacobian matching using it is still more effective compared to simply using Gaussian perturbations.

\begin{algorithm}[t]
\small
\begin{algorithmic}
\caption{Knowledge distillation from deep ensembles with \gls{ods} perturbations}
\label{alg:method}
    \Require Training data $\calD=\{(\bsx_i,\bsy_i)\}_{i=1}^N$.
    \Require Knowledge distillation weight $\alpha$, temperature $\tau$, learning rate $\beta$, \gls{ods} step-size $\eta$.
    \Require Deep ensemble teacher networks $\{\calT_j\}_{j=1}^M$ where $M$ denotes the size of ensembles.
    \State Let $\calS$ be a student network which consists of $M$ subnetworks $\{\calS_j\}_{j=1}^{M}$.
    \State Initialize parameters $\btheta$ of the student network $\calS$.
    \While{not converged}
        \State Sample $(\bsx,\bsy)\sim\calD$, $\bsw \sim \mathrm{Unif}([-1,1])^K$, and a teacher index $r$ uniformly from $\{1,\dots, M\}$.
        \State Sample diversified inputs $\tilde{\bsx} = \bsx + \eta \bvarepsilon_{\text{ODS}}(\bsx,\softmax(\hat\calT_r(\bsx)/\tau),\bsw)$
        \State $\Lce \gets -\sum_{j=1}^{M}\sum_{k=1}^K y^{(k)} \log \calS_j^{(k)}(\bsx)$
        \State $\Lkd \gets -\tau^2\sum_{j=1}^{M}\sum_{k=1}^K \softmax^{(k)}( \hat{\calT}(\tilde{\bsx}) / \tau ) \log\softmax^{(k)}( \hat{\calS}_j(\tilde{\bsx}) / \tau )$
        \State $\btheta \gets \btheta - \beta\nabla_{\btheta}((1-\alpha)\Lce + \alpha\Lkd)$
    \EndWhile
\end{algorithmic}
\end{algorithm}


\section{Related Works}
\label{sec:related_works}

\paragraph{Knowledge distillation}
\Gls{kd} is a method for transferring the information inside a large model into a smaller one \citep{hinton2015distilling} by minimizing the KL divergence between teacher and student networks.
\citep{wang2020knowledge} finds that additionally augmenting the data results in further performance gains because such augmentations allow the \gls{kd} loss to tap into teacher information outside of the training set.
Furthermore, modifications to the standard \gls{kd} objective allow one to distill the knowledge in an ensemble of teachers into one student network while preserving the benefits in uncertainty estimation of ensembles \citep{malinin2019ensemble,tran2020hydra}.
\Gls{kd} can be enhanced via gradient matching, and several existing methods~\citep{srinivas2018jacobian, czarnecki2017sobolev} propose sampling methods for knowledge distillation up to higher-order gradients.
However, such existing methods suffer from a high computational burden or inefficient random perturbation.

\paragraph{Ensembles}
Ensemble methods \citep{hansen90ensemble,dietterich20ensemble}, which construct a set of learners and make predictions on new data points by a weighted average, have been studied extensively.
Ensembles perform at least as well as each individual member \citep{krogh94ensembles}, and achieve the best performance when each member makes errors independently.
Aside from benefits in model accuracy, \gls{de} \citep{lakshminarayanan2017simple,ovadia2019trust} have recently shown to be a simple and scalable alternative to Bayesian neural networks because of their superior uncertainty estimation performance.
\citet{wen2019batchensemble} proposes a network architecture that uses a rank-one parameterization to train an ensemble with low computation and memory cost.


\section{Experiments}
\label{sec:experiments}

In this section, we try to answer the following questions with empirical validation.
\begin{itemize}
    \item Do images perturbed by \gls{ods} increase diversity in \gls{de} teacher predictions? - \cref{subsec:exp_ods_diversity}.
    \item Do \gls{ods} perturbations encourage Jacobian matching? - \cref{subsec:exp_ods_grad}.
    \item Does enhanced diversity of students lead to improved performance in terms of prediction accuracy and uncertainty estimation? - \cref{subsec:exp_main} and \cref{subsec:app_uncertainty}.
\end{itemize}

\subsection{Experimental setup}
\label{subsec:exp_setting}

\paragraph{Datasets and networks} We compared our methods on CIFAR-10/100 and TinyImageNet.
We used ResNet-32~\citep{he2016deep} for CIFAR-10, and WideResNet-28x10~\citep{zagoruyko2016wrn} for CIFAR-100 and TinyImageNet.
Please refer to \cref{subsec:app_training} for detailed training settings.

\paragraph{Hyperparameter settings} The important hyperparameters for \gls{kd} are the pair $(\alpha, \tau)$; for CIFAR-10, after a through hyperparameter sweep, we decided to stay consistent with the convention of $(\alpha, \tau) = (0.9, 4)$ for all methods~\citep{hinton2015distilling,cho2019efficacy,wang2020knowledge}.
For CIFAR-100 and TinyImageNet, we used the value $(\alpha, \tau) = (0.9, 1)$ for all methods.
We fix \gls{ods} step-size $\eta$ to $1/255$ across all settings.
See \cref{subsec:app_training} for more details.

\paragraph{Uncertainty metrics} We measure uncertainty calibration performance with common metrics including \gls{acc}, \gls{nll}, \gls{ece} and \gls{bs}.
We also report the \gls{dee} score~\citep{ashukha2020pitfalls}, which measures the effective number of models by comparing it to a full \gls{de}.
For all methods, the metrics are computed at optimal temperatures obtained from temperature scaling~\citep{guo2017ts}, as suggested by \citet{ashukha2020pitfalls}.
See \cref{subsec:app_evaluation} for the definition of each metric.

\subsection{\texorpdfstring{Impact of \gls{ods} on diversities}{Impact of ODS on diversities}}
\label{subsec:exp_ods_diversity}

\begin{figure}
    \centering
    \begin{subfigure}{\linewidth}
        \includegraphics[width=\linewidth]{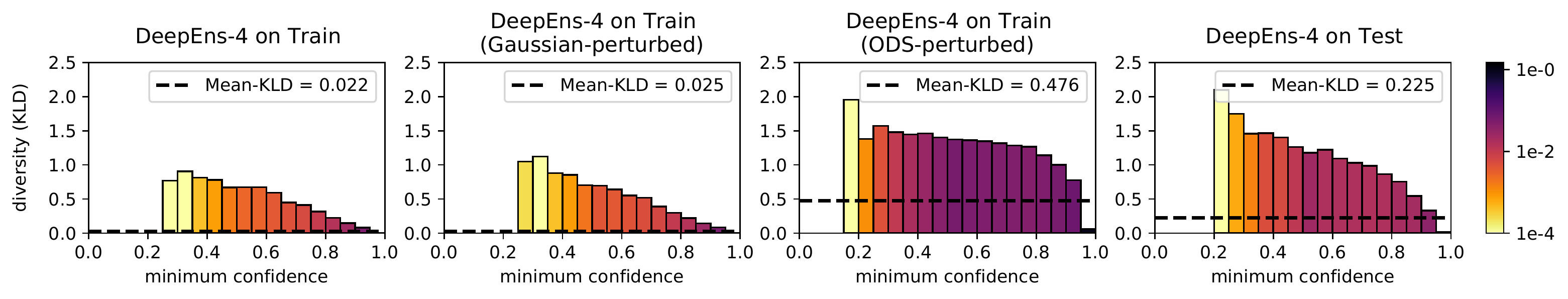}
        \caption{Diversity plots of \gls{de}-4 teachers for ResNet-32 on train examples of CIFAR-10.}
        \label{fig:diversity_de_trn}
    \end{subfigure}
    \begin{subfigure}{\linewidth}
        \vspace{1em}
        \includegraphics[width=\linewidth]{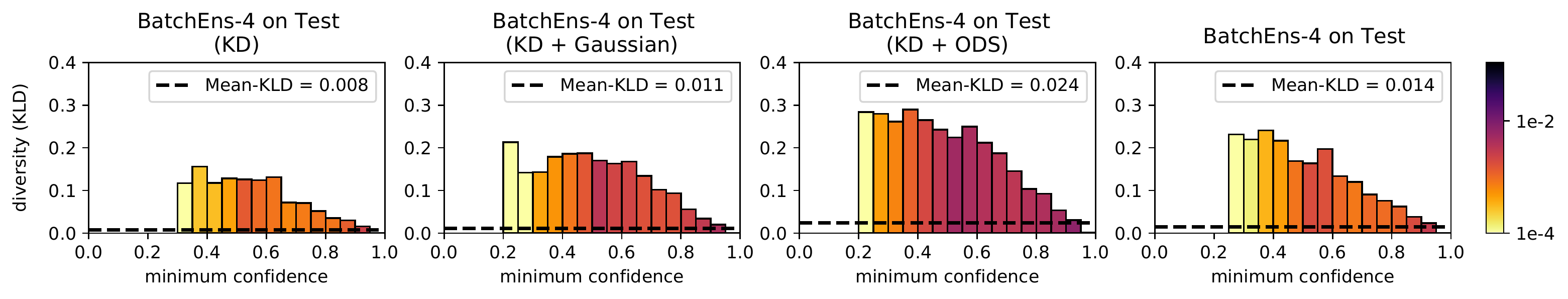}
        \caption{Diversity plots of \gls{be}-4 students for ResNet-32 on test examples of CIFAR-10.}
        \label{fig:diversity_be_tst}
    \end{subfigure}
    \caption{Bar colors denote the density of the bins, \ie, ratio of samples belonging to the bins.}
    \label{fig:diversity_plots}
\end{figure}

In order to assess the effect of \gls{ods} perturbation on the diversities of \gls{de} teachers or student subnetworks, we draw \emph{diversity plots} in \cref{fig:diversity_plots}.
To draw a diversity plot, we first collect the outputs of target models evaluated on a specific dataset and binned the samples w.r.t. their confidence values (minimum confidence among $M$ models).
For each bin, we computed average pairwise KL-divergence between class output probabilities of ensemble members.
We also report the representative value (mean-KLD), which is defined as average KL-divergence values weighted by bin counts.

\cref{fig:diversity_de_trn} shows the diversity plots of \gls{de} teachers on CIFAR-10.
Intuitively, the samples predicted with high confidences usually come with low diversities, and this is indeed depicted in the plots.
For the \gls{de} evaluated on the training set, the mean-KLD value is very small (the first box) compared to the one computed on the test set (the fourth box).
While Gaussian perturbation does not affect the diversities (the second box), \gls{ods} drastically amplifies the diversities (the third box).

\cref{fig:diversity_be_tst} shows the diversity plots of the \gls{be} students trained on CIFAR-10.
Although evaluated at the test set, the \gls{be} distilled without any perturbation (the first box), the \gls{be} distilled with Gaussian perturbations (the second box), and the \gls{be} trained from scratch (the fourth box) exhibit very low diversities.
On the other hand, the \gls{be} distilled with \gls{ods} perturbations show relatively high diversity, and as we will show later in \cref{subsec:exp_main}, this indeed leads to improved performances.

In \cref{sec:methods}, we assumed that an \gls{ods} computed from a specific teacher can be transferred to other teachers.
As an empirical justification, we draw the diversity plot using \gls{ods} computed from an external teacher (a network trained in the same way as the teachers but not actually being used during distillation) in \cref{fig:diversity_external}.
The mean-KLD values are indeed increased compared to the original \gls{de} or \gls{de} evaluated on examples perturbed with Gaussian noise, confirming the validity of our assumption.

\subsection{\texorpdfstring{\gls{ods} for Jacobian matching}{ODS for Jacobian matching}}
\label{subsec:exp_ods_grad}

\glsunset{snr} 
\begin{wrapfigure}{r}{0.35\textwidth}
    \centering
    \vspace{-20pt}
    \includegraphics[width=\linewidth]{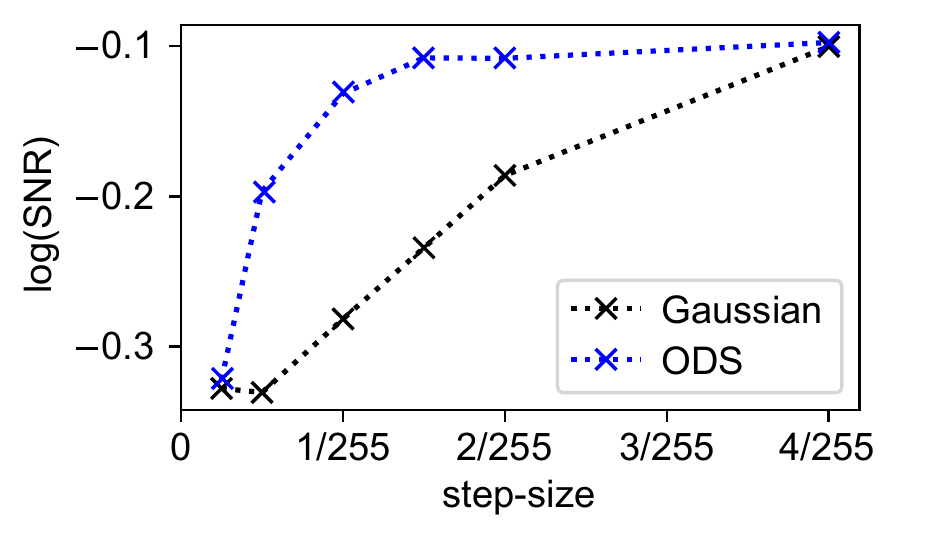}
    \vspace{-20pt}
    \caption{\gls{snr} values computed with Gaussian and \gls{ods} perturbations with varying step-sizes.}
    \label{fig:grad_snr}
    \vspace{-10pt}
\end{wrapfigure}
\glsreset{snr}

\begin{figure}
    \begin{minipage}[c]{0.49\linewidth} \centering
        \begin{subfigure}{0.48\linewidth}
            \includegraphics[height=3.7cm]{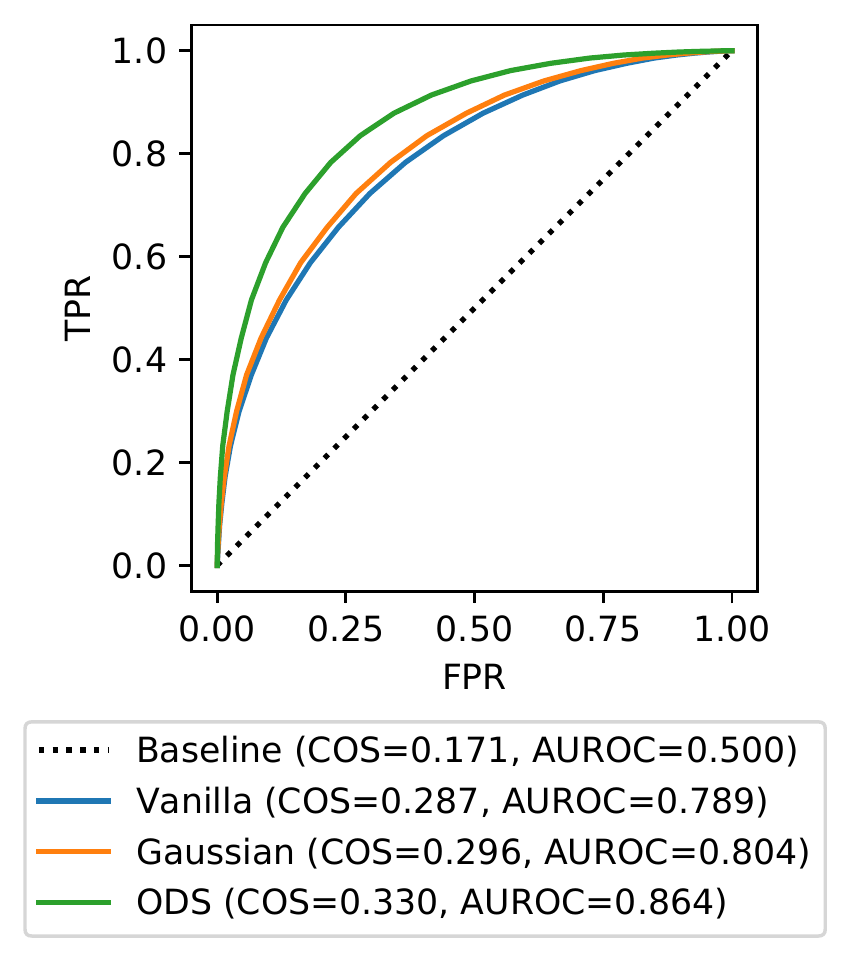}
            \caption{Training set}
            \label{fig:roc_grad_trn}
        \end{subfigure}
        \begin{subfigure}{0.48\linewidth}
            \includegraphics[height=3.7cm]{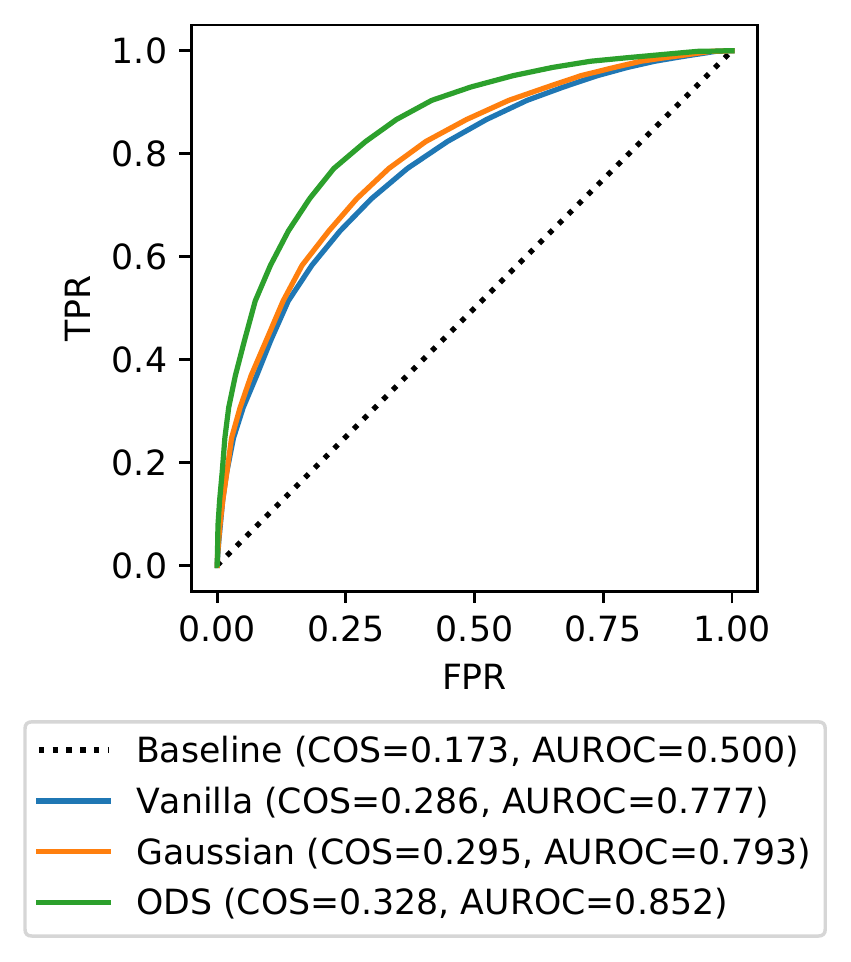}
            \caption{Validation set}
            \label{fig:roc_grad_val}
        \end{subfigure}
    \caption{\gls{roc}-curves with cosine similarities of Jacobians between a teacher and students. `COS' denotes the cosine similarities between the teacher and student averaged over examples, and `AUROC' denotes the area under the \gls{roc}-curves.}
    \label{fig:roc_grad}
    \end{minipage}
    \hfill
    \begin{minipage}[c]{0.49\linewidth} \centering
        \begin{subfigure}{0.48\linewidth}
            \includegraphics[height=3.7cm]{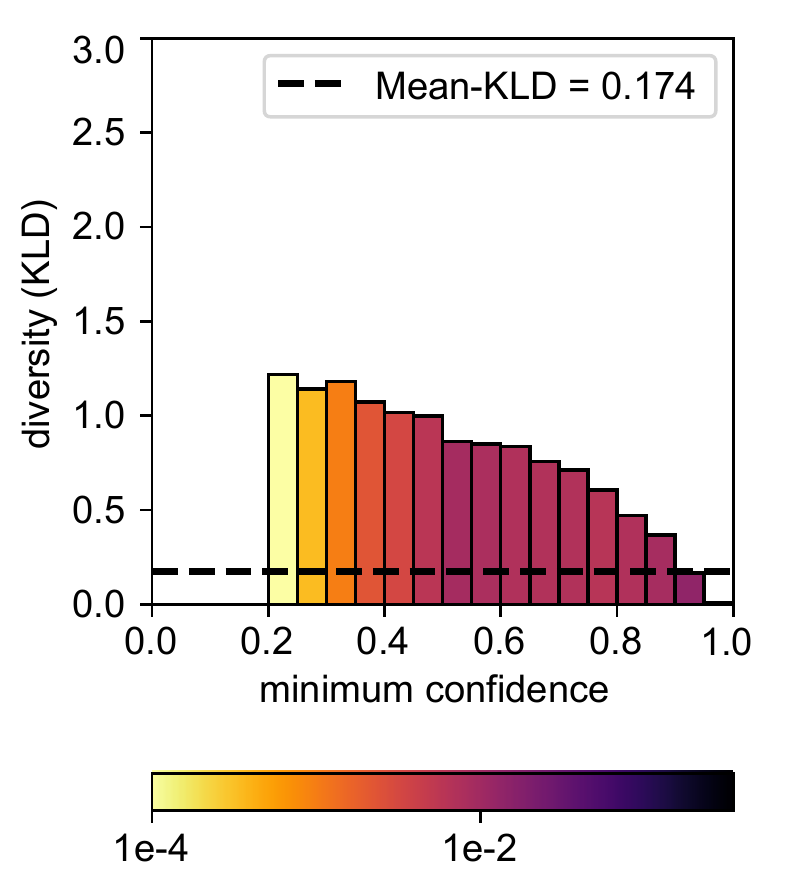}
            \caption{Diversity plot}
            \label{fig:diversity_external}
        \end{subfigure}
        \begin{subfigure}{0.48\linewidth}
            \includegraphics[height=3.7cm]{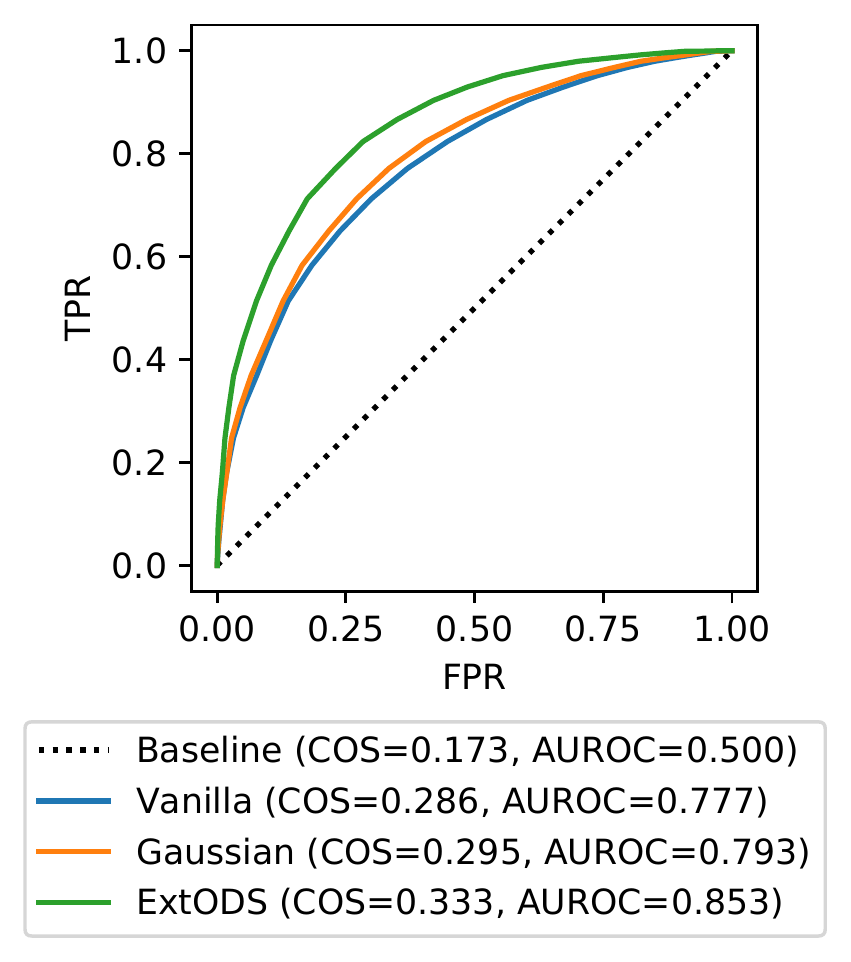}
            \caption{\gls{roc}-curve}
            \label{fig:roc_grad_trf}
        \end{subfigure}
    \caption{Transferability analysis using diversity plot and \gls{roc}-curve. Here, a \gls{be} student is trained with \gls{ods} perturbations computed from an external teacher, and it is denoted as `ExtODS' in the \gls{roc}-curve for validation set.}
    \label{fig:transfer}
    \end{minipage}
\end{figure}

\paragraph{Effectiveness}
We empirically validate our conjecture that \gls{ods} perturbations are more effective in Jacobian matching.
To this end, we computed the cosine similarities between the vectorized Jacobians of ResNet-32 teacher and student networks trained with vanilla \gls{kd}.
We computed the same quantity using student networks distilled with Gaussian perturbations and \gls{ods} perturbations.
We draw \gls{roc} curves to compare the similarities of distilled students and the baseline trained from scratch.
Specifically, we compare two distributions of the cosine similarities between the vectorized Jacobians of a teacher network and a student network: (1) when the student network is trained by \gls{kd}, and (2) when the student network is trained from scratch without any guide from the teacher network.
Results in \cref{fig:roc_grad} clearly show that the Jacobians of the student distilled with \gls{ods} better match the Jacobians, while the student distilled with Gaussian perturbations did not significantly improve upon the student distilled with vanilla \gls{kd}.

Next, we compute the gradient of the term related to the Jacobian matching, \ie,
\[
G(\bsx, \bvarepsilon) = \nabla_{\btheta} ( \Lkd(\calS(\bsx+\bvarepsilon), \calT(\bsx+\bvarepsilon)) - \Lkd(\calS(\bsx), \calT(\bsx)) ).
\]Using both Gaussian perturbations and \gls{ods} perturbations, we collected multiple gradient samples and measured the \gls{snr} of the gradient norms. Specifically, we computed 
$
\operatorname{SNR}(\bsx, \bvarepsilon) =
\norm{\bbE_\bvarepsilon\left[ \operatorname{vec}(G(\bsx,\bvarepsilon))\right]}_2
/
\sqrt{\norm{\operatorname{Var}\left[ \operatorname{vec}(G(\bsx,\bvarepsilon)) \right]}_2}
$,
where $\operatorname{vec}(\cdot)$ is vectorization and $\operatorname{Var}[\cdot]$ is element-wise variance.
As shown in \cref{fig:grad_snr}, the gradients computed with \gls{ods} perturbations exhibit higher \gls{snr}, indicating that learning with such gradients is more stable and efficient.

\paragraph{Transferability}
As in \cref{subsec:exp_ods_diversity}, we performed an experiment to verify our assumption that \gls{ods} perturbations for a specific model are transferable to other models, for the purpose approximate Jacobian matching.
We trained a \gls{be} student with \gls{ods} perturbations computed from an external teacher, and measured the cosine similarities between the student and teachers (\emph{not} between the students and the external teacher).
As one can see from \cref{fig:roc_grad_trf}, the resulting student network copied the Jacobians of the teachers significantly better than the baselines.

\subsection{Main results: image classification tasks}
\label{subsec:exp_main}

\begin{table}[t]
    \caption{Knowledge distillation from \gls{de}-$M$ into \gls{be}-$M$ where $M$ denotes the size of ensembles:
    \gls{acc} and \textit{calibrated metrics} including \gls{nll}, \gls{bs}, \gls{ece}, and \gls{dee}.
    All values for ResNet-32 and WideResNet-28x10 are averaged over four and three experiments, respectively.\\}
    \label{tab:table_kd}
    \centering
    \resizebox{0.95\linewidth}{!}{ 
    \begin{tabular}{lrlllll}
    \toprule
    & & \multicolumn{5}{c}{\textbf{BatchEns << DeepEns (ResNet-32 on CIFAR-10)}} \\
    \cmidrule(r){3-7}
    Method                            &  \# Params & \gls{acc} ($\uparrow$) & \gls{nll} ($\downarrow$) & \gls{bs} ($\downarrow$) & \gls{ece} ($\downarrow$) & \gls{dee} ($\uparrow$) \\
    \midrule
    $\calT$ : DeepEns-4               &   $1.86$ M & $94.42$           & $0.167$            & $0.082$            & $0.008$            & -                  \\
    $\calS$ : BatchEns-4              &   $0.47$ M & $93.37\spm{0.11}$ & $0.204\spm{0.002}$ & $0.099\spm{0.001}$ & $0.008\spm{0.001}$ & $1.419\spm{0.075}$ \\
    \hspace{4mm}+ KD                  &            & $93.98\spm{0.20}$ & $0.188\spm{0.003}$ & $0.091\spm{0.002}$ & $0.009\spm{0.002}$ & $2.019\spm{0.174}$ \\
    \hspace{4mm}+ KD + Gaussian       &            & $93.93\spm{0.12}$ & $0.187\spm{0.001}$ & $0.090\spm{0.001}$ & $0.009\spm{0.002}$ & $2.042\spm{0.089}$ \\
    \tikzmark{startrecc}\hspace{4mm}+ KD + ODS            &            & $93.89\spm{0.10}$ & $0.181\spm{0.002}$ & $0.090\spm{0.001}$ & $\mathbf{0.006}\spm{0.001}$ & $2.486\spm{0.164}$ \\
    \hspace{4mm}+ KD + ConfODS        &            & $\mathbf{94.01}\spm{0.19}$ & $\mathbf{0.180}\spm{0.001}$ & $\mathbf{0.089}\spm{0.001}$ & $0.007\spm{0.001}$ & $\mathbf{2.524}\spm{0.080}$ \tikzmark{endrecc} \\
    \midrule
    $\calT$ : DeepEns-8               &   $3.71$ M & $94.78$           & $0.157$            & $0.077$            & $0.005$            & -                  \\
    $\calS$ : BatchEns-8              &   $0.48$ M & $93.47\spm{0.14}$ & $0.202\spm{0.005}$ & $0.098\spm{0.002}$ & $\mathbf{0.006}\spm{0.002}$ & $1.494\spm{0.156}$ \\
    \hspace{4mm}+ KD                  &            & $94.15\spm{0.13}$ & $0.182\spm{0.001}$ & $0.088\spm{0.001}$ & $0.010\spm{0.001}$ & $2.391\spm{0.113}$ \\
    \hspace{4mm}+ KD + Gaussian       &            & $94.09\spm{0.08}$ & $0.184\spm{0.002}$ & $0.089\spm{0.001}$ & $0.010\spm{0.002}$ & $2.206\spm{0.175}$ \\
    \tikzmark{startrecb}\hspace{4mm}+ KD + ODS            &            & $94.13\spm{0.08}$ & $0.175\spm{0.003}$ & $\mathbf{0.086}\spm{0.001}$  & $\mathbf{0.006}\spm{0.002}$ & $2.991\spm{0.322}$ \\
    \hspace{4mm}+ KD + ConfODS        &            & $\mathbf{94.18}\spm{0.12}$ & $\mathbf{0.174}\spm{0.002}$ & $\mathbf{0.086}\spm{0.001}$ & $0.007\spm{0.001}$ & $\mathbf{3.064}\spm{0.246}$ \tikzmark{endrecb} \\
    \midrule
    \midrule
    & & \multicolumn{5}{c}{\textbf{BatchEns << DeepEns (WRN-28x10 on CIFAR-100)}} \\
    \cmidrule(r){3-7}
    Method                            &  \# Params & \gls{acc} ($\uparrow$) & \gls{nll} ($\downarrow$) & \gls{bs} ($\downarrow$) & \gls{ece} ($\downarrow$) & \gls{dee} ($\uparrow$) \\
    \midrule
    $\calT$ : DeepEns-4               & $146.15$ M & $82.52$           & $0.661$            & $0.247$            & $0.022$            & -                  \\
    $\calS$ : BatchEns-4              &  $36.62$ M & $80.34\spm{0.08}$ & $0.755\spm{0.007}$ & $0.280\spm{0.002}$ & $0.027\spm{0.001}$ & $1.449\spm{0.085}$ \\
    \hspace{4mm}+ KD                  &            & $80.51\spm{0.22}$ & $0.744\spm{0.003}$ & $0.274\spm{0.001}$ & $\mathbf{0.021}\spm{0.003}$ & $1.582\spm{0.035}$ \\
    \hspace{4mm}+ KD + Gaussian       &            & $80.39\spm{0.12}$ & $0.761\spm{0.006}$ & $0.277\spm{0.000}$ & $0.022\spm{0.003}$ & $1.379\spm{0.072}$ \\
    \tikzmark{startreca}\hspace{4mm}+ KD + ODS            &            & $\mathbf{81.88}\spm{0.32}$ & $0.674\spm{0.016}$ & $0.257\spm{0.006}$ & $0.026\spm{0.003}$ & $3.303\spm{0.769}$ \\
    \hspace{4mm}+ KD + ConfODS        &            & $81.85\spm{0.32}$ & $\mathbf{0.672}\spm{0.010}$ & $\mathbf{0.256}\spm{0.003}$ & $0.024\spm{0.000}$ & $\mathbf{3.333}\spm{0.491}$ \tikzmark{endreca} \\
    \bottomrule
    \begin{tikzpicture}[remember picture,overlay]
    \foreach \Val in {reca,recb,recc}
    {
    \draw[rounded corners,black,dotted]
      ([shift={(1.5ex,2ex)}]pic cs:start\Val) 
        rectangle 
      ([shift={(0.5ex,-0.5ex)}]pic cs:end\Val);
    }
    \end{tikzpicture}
    \end{tabular} }
\vspace{-0.5cm}
\end{table}

We validate whether the enhanced diversity of student networks is helpful for actual prediction. 
We compared our methods against three baselines: \gls{be} trained from scratch~\citep{wen2019batchensemble}, \gls{be} distilled without perturbation~\citep{mariet2020bekd}, and \gls{be} distilled with Gaussian noises.

The results are presented in \cref{tab:table_kd}. 
For both datasets, \gls{be} distilled with \gls{ods} or Conf\gls{ods} significantly outperforms baselines in terms of every metric we measured. 
The improvement is much more significant for the larger ensembles ($M=8$ for CIFAR-10) and the larger dataset (CIFAR-100). 
We note that using our diversification strategy improves performance both in terms of raw accuracy and uncertainty metrics.
An especially noteworthy result on CIFAR-100 is that \gls{be} distilled with \gls{ods} achieved a \gls{dee} score over three. 
This is quite remarkable considering the \gls{be} model has far fewer parameters than the full \gls{de} which, by definition, has an expected \gls{dee} of four.
Moreover, further uncertainty results in the \cref{subsec:app_uncertainty} show that our approach is superior with respect to both predictive uncertainty for out-of-distribution examples and calibration on corrupted datasets.

One great benefit of the \gls{kd} framework is that it can transfer knowledge to networks having different architectures, greatly reducing the number of parameters required to achieve a certain level of performance. 
Using CIFAR-100 and TinyImageNet datasets, we compared our method to baselines with student networks having smaller architecture.
The results displayed in \cref{tab:table_kd_cross} show that \gls{kd} with \gls{ods} or Conf\gls{ods} outperforms the baselines by a wide margin.
\begin{table}[t]
    \footnotesize
    \setlength{\tabcolsep}{1em}
    \caption{Cross-architecture knowledge distillation for a model compression on CIFAR-100 and TinyImageNet: \gls{be}-4 students with smaller networks (\ie, WideResNet-28x2 and WideResNet-28x5), learn from the \gls{de}-4 teacher which consists of larger networks (\ie, WideResNet-28x10). All values for CIFAR-100 and TinyImageNet are averaged over three and one experiments, respectively.\\}
    \label{tab:table_kd_cross}
    \centering
    \resizebox{0.95\linewidth}{!}{
    \begin{tabular}{lrllll}
    \toprule
    & & \multicolumn{4}{c}{\textbf{BatchEns-4 << DeepEns-4 (WRN-28x2 on CIFAR-100)}} \\
    \cmidrule(r){3-6}
    Method                         &  \# Params & \gls{acc} ($\uparrow$) & \gls{nll} ($\downarrow$) & \gls{bs} ($\downarrow$) & \gls{ece} ($\downarrow$) \\
    \midrule
    $\calT$ : DeepEns-4            & $146.15$ M & $82.52$           & $0.676$            & $0.250$            & $0.035$ \\
    $\calS$ : BatchEns-4           &   $1.50$ M & $75.17\spm{0.27}$ & $1.245\spm{0.024}$ & $0.383\spm{0.004}$ & $0.141\spm{0.004}$ \\
    \hspace{4mm}+ KD               &            & $75.19\spm{0.36}$ & $1.207\spm{0.021}$ & $0.377\spm{0.007}$ & $0.136\spm{0.005}$ \\
    \hspace{4mm}+ KD + Gaussian    &            & $74.50\spm{0.17}$ & $1.247\spm{0.012}$ & $0.389\spm{0.004}$ & $0.137\spm{0.003}$ \\
    \tikzmark{startrecf}\hspace{4mm}+ KD + ODS         &            & $\mathbf{76.03}\spm{0.22}$ & $\mathbf{0.899}\spm{0.013}$ & $\mathbf{0.333}\spm{0.003}$ & $\mathbf{0.027}\spm{0.002}$ \\
    \hspace{4mm}+ KD + ConfODS     &            & $76.01\spm{0.16}$ & $0.901\spm{0.006}$ & $0.334\spm{0.002}$ & $0.028\spm{0.004}$ \tikzmark{endrecf}\\
    \midrule
    & & \multicolumn{4}{c}{\textbf{BatchEns-4 << DeepEns-4 (WRN-28x5 on CIFAR-100)}} \\
    \cmidrule(r){3-6}
    Method                         &  \# Params & \gls{acc} ($\uparrow$) & \gls{nll} ($\downarrow$) & \gls{bs} ($\downarrow$) & \gls{ece} ($\downarrow$) \\
    \midrule
    $\calT$ : DeepEns-4            & $146.15$ M & $82.52$           & $0.661$            & $0.247$            & $0.022$ \\
    $\calS$ : BatchEns-4           &   $9.20$ M & $78.75\spm{0.11}$ & $0.801\spm{0.012}$ & $0.297\spm{0.003}$ & $0.021\spm{0.002}$ \\
    \hspace{4mm}+ KD               &            & $78.89\spm{0.10}$ & $0.804\spm{0.012}$ & $0.296\spm{0.003}$ & $0.022\spm{0.001}$ \\
    \hspace{4mm}+ KD + Gaussian    &            & $78.80\spm{0.41}$ & $0.815\spm{0.009}$ & $0.297\spm{0.005}$ & $\mathbf{0.020}\spm{0.002}$ \\
    \tikzmark{startrece}\hspace{4mm}+ KD + ODS         &            & $80.24\spm{0.05}$ & $0.742\spm{0.008}$ & $0.279\spm{0.002}$ & $0.028\spm{0.004}$ \\
    \hspace{4mm}+ KD + ConfODS     &            & $\mathbf{80.62}\spm{0.25}$ & $\mathbf{0.733}\spm{0.007}$ & $\mathbf{0.275}\spm{0.003}$ & $0.027\spm{0.001}$ \tikzmark{endrece}\\
    \midrule
    \midrule
    & & \multicolumn{4}{c}{\textbf{BatchEns-4 << DeepEns-4 (WRN-28x5 on TinyImageNet)}} \\
    \cmidrule(r){3-6}
    Method                         &  \# Params & \gls{acc} ($\uparrow$) & \gls{nll} ($\downarrow$) & \gls{bs} ($\downarrow$) & \gls{ece} ($\downarrow$) \\
    \midrule
    $\calT$ : DeepEns-4            & $146.40$ M & $69.90$ & $1.242$ & $0.403$ & $0.016$ \\
    $\calS$ : BatchEns-4           &   $9.23$ M & $64.86$ & $1.455$ & $0.464$ & $\mathbf{0.022}$ \\
    \hspace{4mm}+ KD               &            & $65.86$ & $1.432$ & $\mathbf{0.456}$ & $0.022$ \\
    \hspace{4mm}+ KD + Gaussian    &            & $65.72$ & $1.446$ & $0.457$ & $0.022$ \\
    \tikzmark{startrecd}\hspace{4mm}+ KD + ODS         &            & $\mathbf{65.98}$ & $\mathbf{1.408}$ & $\mathbf{0.456}$ & $0.022\color{white}\spm{0.000}$ \tikzmark{endrecd}\\
    \bottomrule
    \begin{tikzpicture}[remember picture,overlay]
    \foreach \Val in {recd,rece,recf}
    {
    \draw[rounded corners,black,dotted]
      ([shift={(1.5ex,2ex)}]pic cs:start\Val) 
        rectangle 
      ([shift={(0.5ex,-0.5ex)}]pic cs:end\Val);
    }
    \end{tikzpicture}
    \end{tabular}}
\vspace{-0.2cm}
\end{table}


\section{Conclusion}
\label{sec:conclusion}

In this paper, we proposed a simple yet effective training scheme for distilling knowledge from an ensemble of teachers.
The key idea of our approach is to perturb training data in a proper way to expose student networks to the diverse outputs produced by ensemble teachers.
We employed \gls{ods} to find such diversifying perturbation. Our distillation scheme with \gls{ods} perturbation can be interpreted as a sample-efficient approximate Jacobian matching procedure.
We empirically verified that our method enhances the diversities of students, and that such enhanced diversities help to reduce the performance gap between student and teacher models in \gls{kd}.
An interesting future work would be to rigorously analyze the Jacobian matching aspect of our method and to provide theoretical justification and a better distillation method to harness it.
Also, in this paper, we constrained ourselves to \gls{be} students, and applying our framework to other student network architectures such as Bayesian neural networks would be interesting.

\paragraph{Limitations}
Although we have shown promising results, our method still has several limitations.
Due to its nature as a \gls{kd} framework, the need to train $M$ teachers involves a considerable computational cost.
Also, our method cannot improve the student beyond its capacity.
For instance, in \cref{tab:table_kd_cross}, WideResNet-28x2 distilled from WideResNet-28x10 does not greatly benefit from our method.
Arguably the most critical limitation is that our method is a heuristic and lacks any theoretical guarantee.
In that sense, as mentioned above, rigorously analyzing the effect of \gls{ods} perturbation will be important future research.


\clearpage

\section*{Acknowledgement}
This work was partly supported by Institute of Information \& communications Technology Planning \& Evaluation (IITP) grant funded by the Korea government (MSIT) (No.2019-0-00075, Artificial Intelligence Graduate School Program(KAIST), No. 2021-0-02068, Artificial Intelligence Innovation Hub),  National Research Foundation of Korea (NRF) funded by the Ministry of Education (NRF-2021R1F1A1061655, NRF-2021M3E5D9025030), and  Samsung Electronics Co., Ltd (IO201214-08176-01). 

\bibliography{references}


\clearpage

\appendix
\suppressfloats


\section{Additional Experiments}
\label{sec:app_additional_experiments}

\subsection{Further evaluation on uncertainty}
\label{subsec:app_uncertainty}

\begin{figure}
    \centering
    \begin{subfigure}{\linewidth}
        \includegraphics[width=\linewidth]{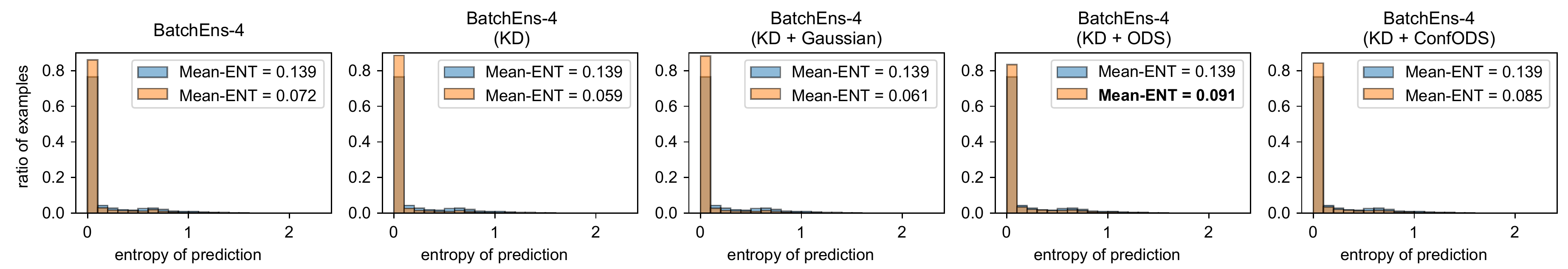}
    \end{subfigure}
    \begin{subfigure}{\linewidth}
        \includegraphics[width=\linewidth]{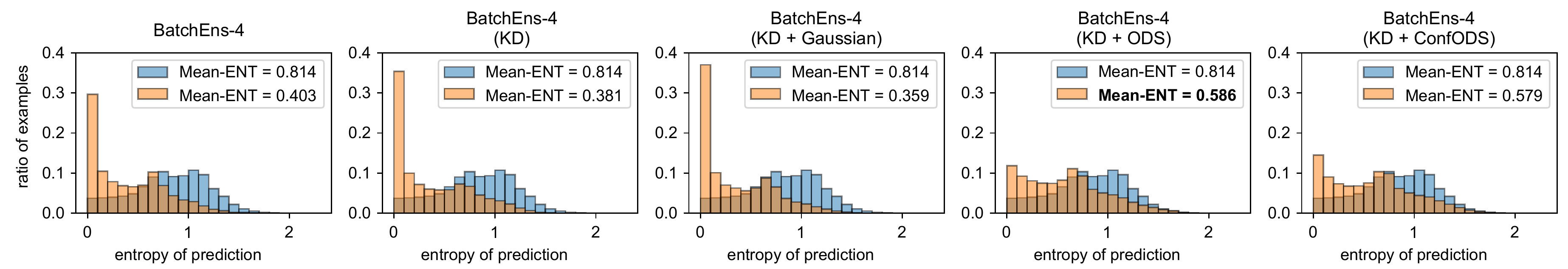}
    \end{subfigure}
    \caption{Histogram of the predictive entropy of ResNet-32 models on test examples from known classes, \ie, CIFAR-10 (top row), and unknown classes, \ie, SVHN (bottom row).
    The histograms with mean entropies of $0.139$ on CIFAR-10 and $0.814$ on SVHN denote DeepEns-4 teacher.}
    \label{fig:plot_ent_c10}
    \vspace{1em}
    \begin{subfigure}{\linewidth}
        \includegraphics[width=\linewidth]{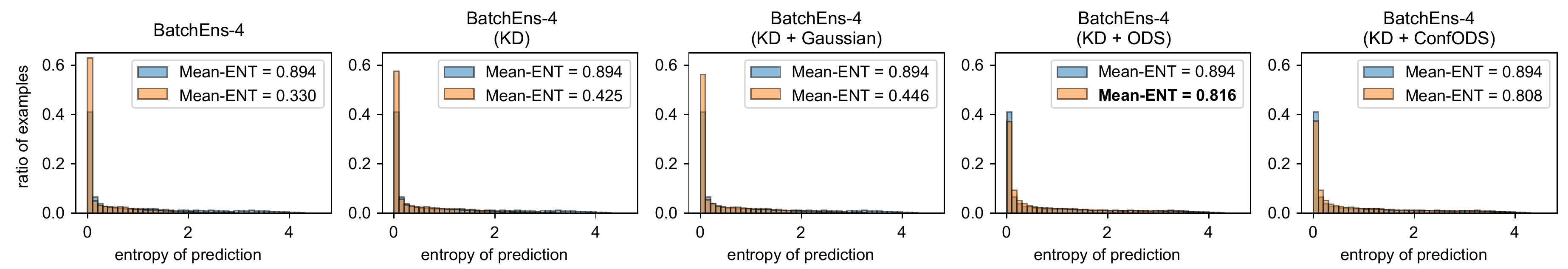}
    \end{subfigure}
    \begin{subfigure}{\linewidth}
        \includegraphics[width=\linewidth]{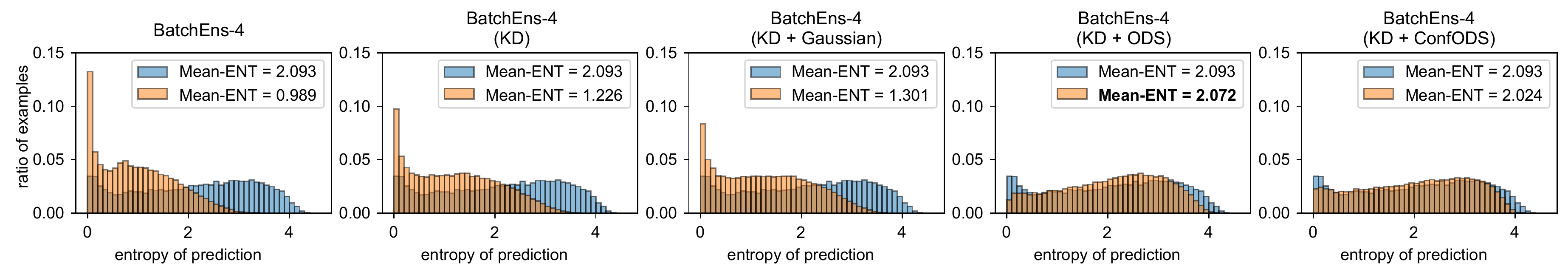}
    \end{subfigure}
    \caption{Histogram of the predictive entropy of WideResNet-28x10 models on test examples from known classes, \ie, CIFAR-100 (top row), and unknown classes, \ie, SVHN (bottom row).
    The histograms with mean entropies of $0.894$ on CIFAR-100 and $2.093$ on SVHN denote DeepEns-4 teacher.}
    \label{fig:plot_ent_c100}
\end{figure}

\paragraph{Predictive uncertainty for out-of-distribution examples}
For reliable deployment in real-world decision making systems, deep neural networks should not make over-confident predictions when inputs are far from the training data.
Following \citet{lakshminarayanan2017simple}, we evaluate the entropy of predictions on out-of-distribution examples from unseen classes to quantify the quality of the predictive uncertainty.
We use the test split of the SVHN dataset as out-of-distribution data; it consists of 26,032 digit images with the same image size as CIFAR-10/100\footnote{\href{http://ufldl.stanford.edu/housenumbers/}{http://ufldl.stanford.edu/housenumbers/}}.
As shown in \cref{fig:plot_ent_c10} and \cref{fig:plot_ent_c100}, while \gls{de} exhibits the highest out-of-distribution entropy, our models show comparable out-of-distribution entropy.
Especially, for WideResNet-28x10 models trained on CIFAR-100, our method almost matches the predictive uncertainty of \gls{de} teachers (\ie, $2.072$ versus $2.093$).

\begin{figure}
    \centering
    \begin{subfigure}{\linewidth}
        \includegraphics[width=\linewidth]{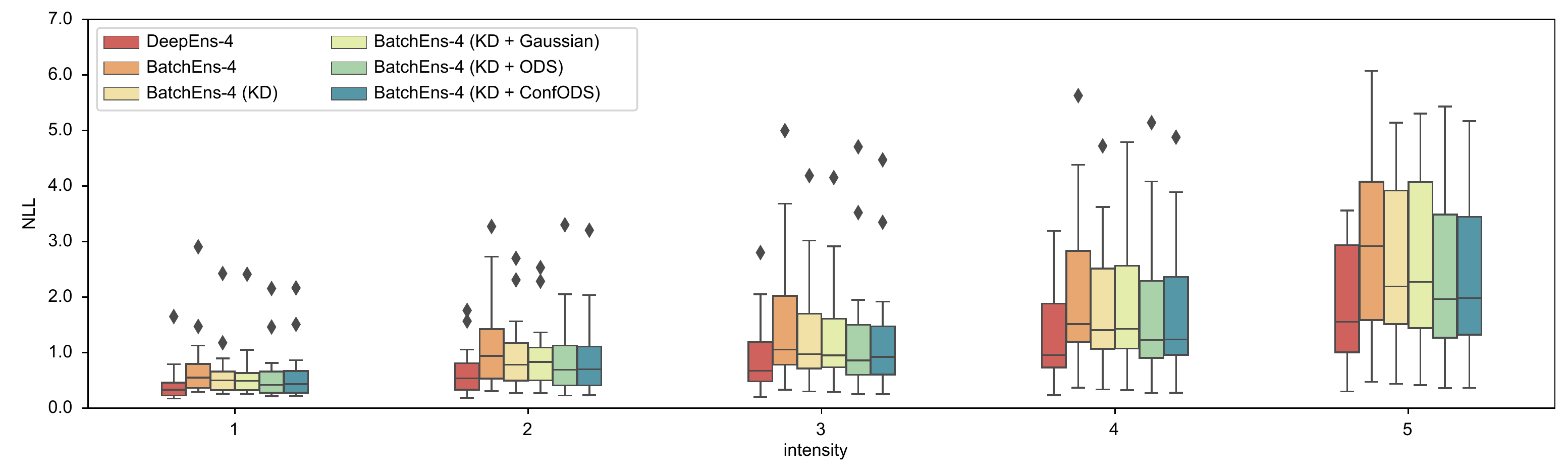}
    \end{subfigure}
    \begin{subfigure}{\linewidth}
        \includegraphics[width=\linewidth]{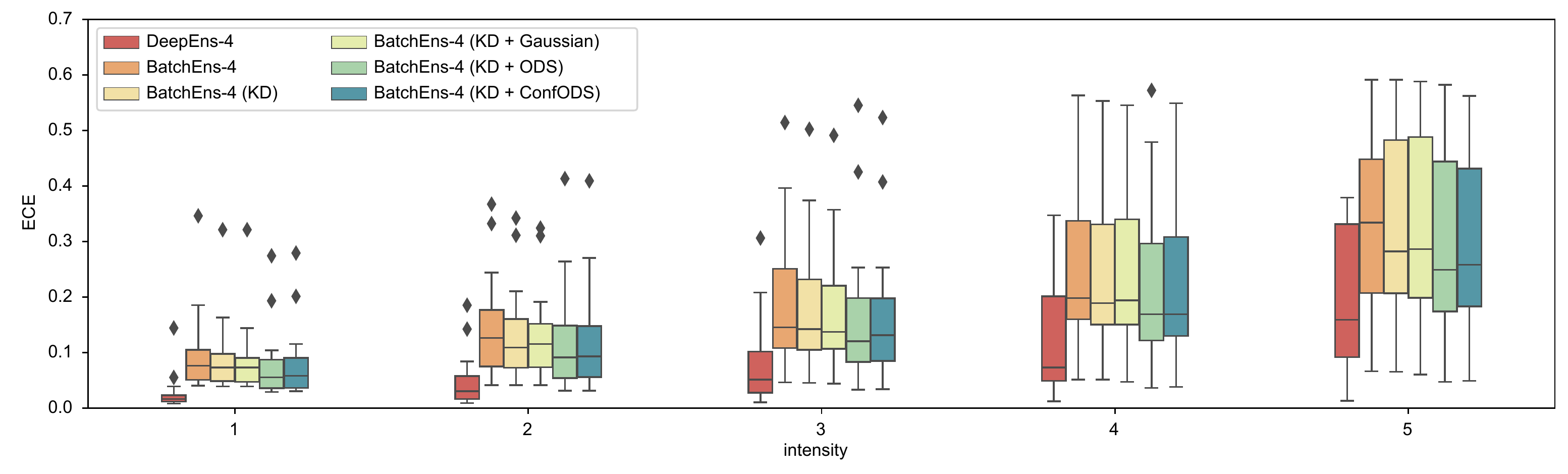}
        \caption{ResNet-32 models on CIFAR-10-C.}
        \vspace{2em}
    \end{subfigure}
    \begin{subfigure}{\linewidth}
        \includegraphics[width=\linewidth]{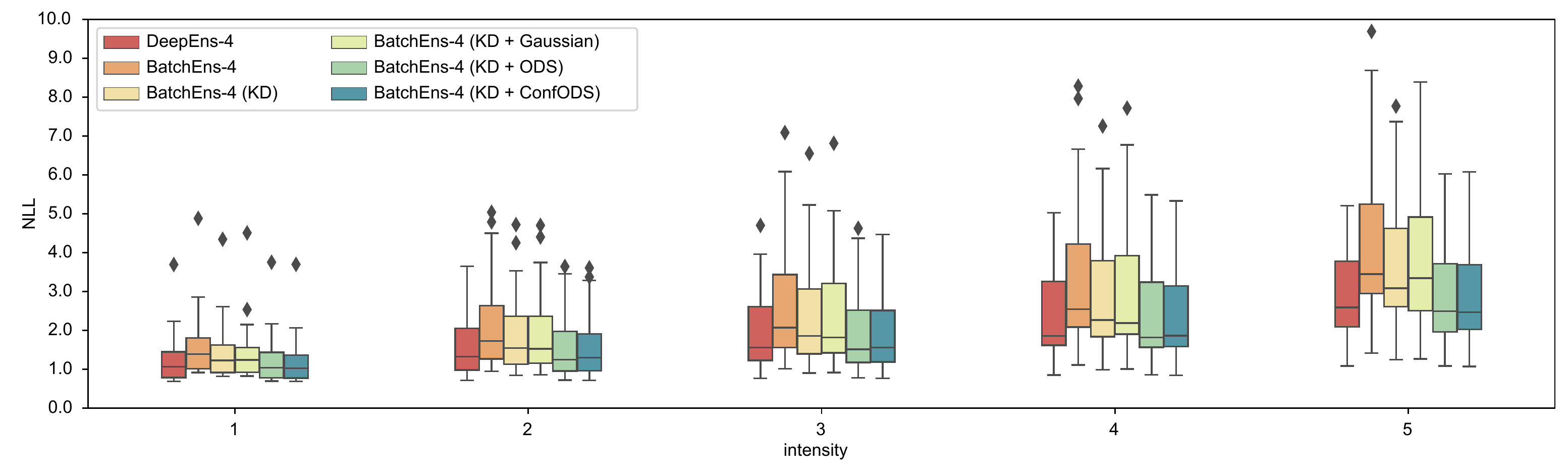}
    \end{subfigure}
    \begin{subfigure}{\linewidth}
        \includegraphics[width=\linewidth]{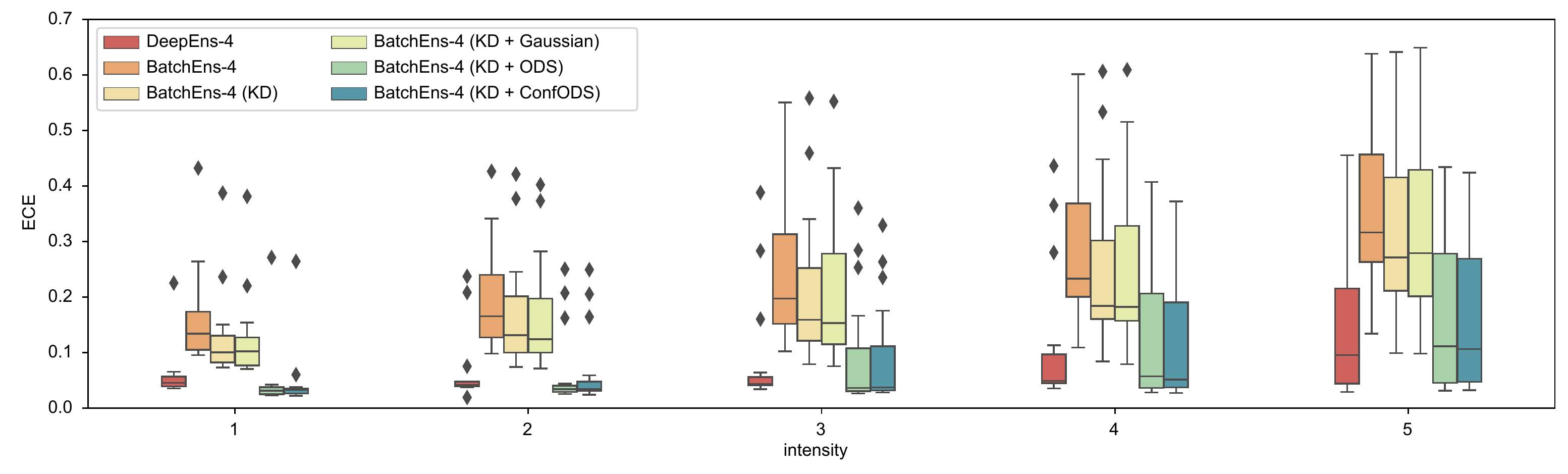}
        \caption{WideResNet-28x10 models on CIFAR-100-C.}
        \vspace{2em}
    \end{subfigure}
    \caption{Calibration on CIFAR-10-C and CIFAR-100-C: box-and-whisker plot shows medians of \gls{nll} and \gls{ece} across corruption types for five levels of intensity.}
    \label{fig:cifar_corrupted}
\end{figure}

\paragraph{Calibration on corrupted datasets}
We also evaluate predictive uncertainty on the corrupted versions of CIFAR datasets; it consists of 15 types of corruptions on the original test examples of CIFAR datasets where each corruption type has five intensities~\citep{hendrycks2019benchmarking}.
In order to compare models in terms of calibration on corrupted datasets, we draw box-and-whisker plots for \gls{nll} and \gls{ece} measuring calibration across corruption types and intensities in \cref{fig:cifar_corrupted}.
It clearly shows that \gls{be} students distilled with \gls{ods} are better calibrated than the baselines.
We also include standard metrics averaged over all corruption types and intensities in \cref{tab:cifar_corrupted}.
A remarkable result on CIFAR-100 is that \gls{be} students with \gls{ods} achieve comparable calibration to \gls{de} teachers both in \cref{fig:cifar_corrupted} and \cref{tab:cifar_corrupted}.

\begin{table}[t]
    \footnotesize
    \setlength{\tabcolsep}{0.5em}
    \caption{Evaluation results on CIFAR-10-C and CIFAR-100-C: \gls{acc} and \textit{standard metrics} including \gls{nll}, \gls{bs}, and \gls{ece}.
    All values for ResNet-32 and WideResNet-28x10 are averaged over four and three experiments, respectively.\\}
    \label{tab:cifar_corrupted}
    \centering
    \begin{tabular}{lrllll}
    \toprule
    & & \multicolumn{4}{c}{\textbf{BatchEns-4 << DeepEns-4 (ResNet-32 on CIFAR-10)}} \\
    \cmidrule(r){3-6}
    Method                         &  \# Params & \gls{acc} ($\uparrow$) & \gls{nll} ($\downarrow$) & \gls{bs} ($\downarrow$) & \gls{ece} ($\downarrow$) \\
    \midrule
    $\calT$ : DeepEns-4            &   $1.86$ M & $73.51$           & $1.036$            & $0.380$            & $0.093$ \\
    $\calS$ : BatchEns-4           &   $0.47$ M & $70.39\spm{0.65}$ & $1.723\spm{0.086}$ & $0.485\spm{0.012}$ & $0.204\spm{0.006}$ \\
    \hspace{4mm}+ KD               &            & $72.68\spm{0.49}$ & $1.485\spm{0.051}$ & $0.455\spm{0.010}$ & $0.196\spm{0.006}$ \\
    \hspace{4mm}+ KD + Gaussian    &            & $\mathbf{72.71}\spm{0.38}$ & $1.487\spm{0.026}$ & $\mathbf{0.453}\spm{0.008}$ & $0.194\spm{0.004}$ \\
    \tikzmark{startrecaaa}\hspace{4mm}+ KD + ODS         &            & $70.50\spm{0.46}$ & $1.406\spm{0.044}$ & $0.462\spm{0.009}$ & $\mathbf{0.178}\spm{0.006}$ \\
    \hspace{4mm}+ KD + ConfODS     &            & $70.80\spm{0.36}$ & $\mathbf{1.404}\spm{0.038}$ & $0.459\spm{0.008}$ & $0.180\spm{0.005}$ \tikzmark{endrecaaa}\\
    \midrule
    & & \multicolumn{4}{c}{\textbf{BatchEns-4 << DeepEns-4 (WRN28x10 on CIFAR-100)}} \\
    \cmidrule(r){3-6}
    Method                         &  \# Params & \gls{acc} ($\uparrow$) & \gls{nll} ($\downarrow$) & \gls{bs} ($\downarrow$) & \gls{ece} ($\downarrow$) \\
    \midrule
    $\calT$ : DeepEns-4            & $146.15$ M & $53.86$           & $2.091$            & $0.609$            & $0.096$ \\
    $\calS$ : BatchEns-4           &  $36.62$ M & $52.36\spm{0.44}$ & $2.953\spm{0.124}$ & $0.708\spm{0.009}$ & $0.255\spm{0.006}$ \\
    \hspace{4mm}+ KD               &            & $53.00\spm{0.33}$ & $2.585\spm{0.023}$ & $0.677\spm{0.005}$ & $0.216\spm{0.003}$ \\
    \hspace{4mm}+ KD + Gaussian    &            & $52.81\spm{0.65}$ & $2.654\spm{0.009}$ & $0.682\spm{0.004}$ & $0.218\spm{0.001}$ \\
    \tikzmark{startrecbbb}\hspace{4mm}+ KD + ODS         &            & $53.87\spm{1.00}$ & $2.109\spm{0.073}$ & $0.614\spm{0.013}$ & $0.104\spm{0.009}$ \\
    \hspace{4mm}+ KD + ConfODS     &            & $\mathbf{54.19}\spm{0.75}$ & $\mathbf{2.083}\spm{0.073}$ & $\mathbf{0.610}\spm{0.015}$ & $\mathbf{0.101}\spm{0.011}$ \tikzmark{endrecbbb}\\
    \bottomrule
    \begin{tikzpicture}[remember picture,overlay]
    \foreach \Val in {recaaa,recbbb}
    {
    \draw[rounded corners,black,dotted]
      ([shift={(1.5ex,2ex)}]pic cs:start\Val) 
        rectangle 
      ([shift={(0.5ex,-0.5ex)}]pic cs:end\Val);
    }
    \end{tikzpicture}
    \end{tabular}
\end{table}

\subsection{\texorpdfstring{Adversarial perturbation instead of \gls{ods}}{Adversarial perturbation instead of ODS}}
\label{subsec:app_adversarial}

The \gls{ods} perturbation is very similar to the adversarial perturbation; if we replace the uniform vector with the one-hot class labels in \cref{eq:ods_def}, we get an adversarial perturbation.
The difference is that the adversarial perturbation is meant to worsen the predictive performance by design because it takes a step toward the directions increasing the classification loss.
We empirically found that the perturbations just increasing diversity without maintaining prediction accuracy can actually harm the performance of student models (this is also related to the performance gain of Conf\gls{ods}).
In specific, we tested the KD with adversarial perturbations (\ie, non-targeted attack) instead of \gls{ods} perturbations using ResNet-32 on CIFAR-10.
This achieved ACC of $93.32\pm0.22$ and NLL of $0.194\pm0.003$, which is much worse than vanilla \gls{kd} with no perturbation presented in \cref{tab:table_kd}.

\subsection{Changing student's architecture}

We performed additional experiments using the \gls{mimo} architecture~\citep{havasi2021mimo}.
The biggest difference between \gls{mimo} and \gls{be} is in how subnetworks are constructed.
In short, unlike \gls{be}, \gls{mimo} constructs its subnetworks without any explicit weight parameterization.
Even though \gls{mimo} constructs its multiple subnetworks implicitly, it can still be trained using our method.
We applied \gls{kd} from \gls{de}-3 teachers to \gls{mimo}-3 students with Gaussian and \gls{ods} perturbations using WideResNet-28x10 on CIFAR-100, and the results are summarized in \cref{tab:table_mimo}.
Again, ours significantly improved the performance compared to the vanilla \gls{kd} or \gls{kd} with Gaussian perturbation.
This result demonstrates that our proposed method is broadly beneficial independently of multi-network architecture and is not specific to \gls{be}.

\begin{table}[t]
    \footnotesize
    \setlength{\tabcolsep}{0.7em}
    \caption{Knowledge distillation from \gls{de}-3 into \gls{mimo}-3: \gls{acc}, \textit{standard metrics} and \textit{calibrated metrics}. All values are measured one time.\\}
    \label{tab:table_mimo}
    \centering
    \resizebox{0.95\linewidth}{!}{ 
    \begin{tabular}{lrccccccc}
    \toprule
    & & \multicolumn{4}{c}{\textbf{Standard Metrics}} & \multicolumn{3}{c}{\textbf{Calibrated Metrics}} \\
    \cmidrule(lr){3-6}\cmidrule(lr){7-9}
    Method                      &  \# Params & \gls{acc} ($\uparrow$) & \gls{nll} ($\downarrow$) & \gls{bs} ($\downarrow$) & \gls{ece} ($\downarrow$) & \gls{nll} ($\downarrow$) & \gls{bs} ($\downarrow$) & \gls{ece} ($\downarrow$) \\
    \midrule
    $\calT$ : DeepEns-3         & $109.61$ M & $82.43$ & $0.690$ & $0.253$ & $0.036$ & $0.677$ & $0.251$ & $0.025$ \\
    $\calS$ : MIMO-3            &  $36.67$ M & $80.63$ & $0.720$ & $0.271$ & $0.027$ & $0.716$ & $0.270$ & $\mathbf{0.014}$ \\
    \hspace{4mm}+ KD            &            & $80.75$ & $0.723$ & $0.271$ & $0.029$ & $0.720$ & $0.271$ & $0.021$ \\
    \hspace{4mm}+ KD + Gaussian &            & $80.69$ & $0.726$ & $0.272$ & $\mathbf{0.024}$ & $0.723$ & $0.271$ & $0.016$ \\
    \tikzmark{startrecccc}\hspace{4mm}+ KD + ODS      &            & $\mathbf{81.17}$ & $\mathbf{0.715}$ & $\mathbf{0.270}$ & $0.037$ & $\mathbf{0.692}$ & $\mathbf{0.267}$ & $0.020$ \tikzmark{endrecccc}\\
    \bottomrule
    \begin{tikzpicture}[remember picture,overlay]
    \foreach \Val in {recccc}
    {
    \draw[rounded corners,black,dotted]
      ([shift={(1.5ex,2ex)}]pic cs:start\Val) 
        rectangle 
      ([shift={(0.5ex,-0.5ex)}]pic cs:end\Val);
    }
    \end{tikzpicture}
    \end{tabular}}
\end{table}

\subsection{\texorpdfstring{Comparison with \gls{endd}}{Comparison with Ensemble Distribution Distillation (END2)}}

To address the issue of distilling diversity more thoroughly, we performed additional experiments comparing to the recently proposed Ensemble Distribution Distillation (\gls{endd})~\citep{malinin2019ensemble}, which considers the distribution of \gls{de} teacher predictions.
Here, we adopted the publicly available PyTorch implementation of \gls{endd}, and compared it with our method using ResNet-32 on CIFAR-10%
\footnote{\href{https://github.com/lennelov/endd-reproduce/tree/d61d298b52c4338e07d7cd4a3fdc65f1de1bcbf1}{https://github.com/lennelov/endd-reproduce/tree/d61d298b52c4338e07d7cd4a3fdc65f1de1bcbf1}}.
Quantitative results in \cref{tab:table_endd} show that ours significantly outperforms \gls{endd} in terms of both accuracy and uncertainty estimation metrics.

\begin{table}[t]
    \footnotesize
    \setlength{\tabcolsep}{0.7em}
    \caption{Comparison between \gls{endd} and ours (\ie, Conf\gls{ods}) distilling from \gls{de}-$M$ teachers where $M$ denotes the size of ensembles using ResNet-32 on CIFAR-10: \gls{acc} and \textit{calibrated metrics} including \gls{nll}, \gls{bs}, and \gls{ece}. All values are measured one time.\\}
    \label{tab:table_endd}
    \centering
    \resizebox{0.95\linewidth}{!}{
    \begin{tabular}{lcccccccc}
    \toprule
    & \multicolumn{4}{c}{\gls{endd}} & \multicolumn{4}{c}{Ours (Conf\gls{ods})} \\
    \cmidrule(lr){2-5}\cmidrule(lr){6-9}
    Teacher              & \gls{acc} ($\uparrow$) & \gls{nll} ($\downarrow$) & \gls{bs} ($\downarrow$) & \gls{ece} ($\downarrow$) & \gls{acc} ($\uparrow$) & \gls{nll} ($\downarrow$) & \gls{bs} ($\downarrow$) & \gls{ece} ($\downarrow$) \\
    \midrule
    $\calT$ : DeepEns-4  & $92.10$ & $0.260$ & $0.121$ & $0.017$ & \tikzmark{startrecaaaaa}$94.01$ & $0.180$ & $0.093$ & $\mathbf{0.007}$ \\
    $\calT$ : DeepEns-8  & $93.11$ & $0.225$ & $0.104$ & $0.013$ & $\mathbf{94.18}$ & $\mathbf{0.174}$ & $\mathbf{0.091}$ & $\mathbf{0.007}$ \tikzmark{endrecaaaaa}\\
    $\calT$ : DeepEns-16 & $93.21$ & $0.221$ & $0.104$ & $0.016$ & -       & -       & -       & -       \\
    $\calT$ : DeepEns-32 & $93.49$ & $0.218$ & $0.102$ & $0.014$ & -       & -       & -       & -       \\
    \bottomrule
    \begin{tikzpicture}[remember picture,overlay]
    \foreach \Val in {recaaaaa}
    {
    \draw[rounded corners,black,dotted]
      ([shift={(-1.0ex,2ex)}]pic cs:start\Val) 
        rectangle 
      ([shift={(0.5ex,-0.5ex)}]pic cs:end\Val);
    }
    \end{tikzpicture}
    \end{tabular}}
\end{table}


\section{Experimental Details}

Code is available at \href{https://github.com/cs-giung/giung2/tree/main/projects/Diversity-Matters}{https://github.com/cs-giung/giung2/tree/main/projects/Diver\-sity-Matters}.

\subsection{Training}
\label{subsec:app_training}

\paragraph{CIFAR-10/100}
CIFAR-10/100 consists of a train set of 50,000 images and a test set of 10,000 images from 10/100 classes, with images size of $32\times32\times3$\footnote{\href{https://www.cs.toronto.edu/~kriz/cifar.html}{https://www.cs.toronto.edu/ \textasciitilde kriz/cifar.html}}.
All models are trained on the first 45k examples of the train split of CIFAR datasets and the last 5k examples of the train split are used as the validation split.
We follow the standard data augmentation policy~\citep{he2016deep} which consists of random cropping of 32 pixels with a padding of 4 pixels and random horizontal flipping.

\paragraph{TinyImageNet}
TinyImageNet is a subset of ImageNet dataset consisting of 100,000 images from 200 classes with images resized to $64\times64\times3$\footnote{\href{http://cs231n.stanford.edu/tiny-imagenet-200.zip}{http://cs231n.stanford.edu/tiny-imagenet-200.zip}}.
Since the labels of the official test set are not publicly available, we use the official validation set consisting of 10,000 images as a test set for experiments.
All models are trained on the first 450 examples for each class and the last 50 examples for each class are used as validation examples, \ie, train and validation splits consist of 90k and 10k examples, respectively.
We use a data augmentation which consists of random cropping of 64 pixels with a padding of 8 pixels and random horizontal flipping.

\paragraph{Learning rate schedules}
We use the following learning rate scheduling suggested in \citet{ashukha2020pitfalls} for all experiments.
\begin{enumerate}
    \item For the first five epochs, linearly increase the learning rate from $0.01\times\texttt{base\_lr}$ to $\texttt{base\_lr}$.
    \item Until $0.5 \times \texttt{total\_epochs}$, keep the learning rate as $\texttt{base\_lr}$.
    \item From $0.5 \times \texttt{total\_epochs}$ to $0.9 \times \texttt{total\_epochs}$, linearly decay the learning rate to $0.01\times\texttt{base\_lr}$.
    \item From $0.9 \times \texttt{total\_epochs}$ to $\texttt{total\_epochs}$, keep the learning rate as $0.01 \times \texttt{base\_lr}$, and save the model with the best validation accuracy.
\end{enumerate}

\paragraph{Optimization}
We use SGD optimizer with momentum $0.9$ for all experiments. Specifically,
\begin{itemize}
    \item We use the optimizer with batch size $512$, \texttt{base\_lr} $0.4$ and weight decay parameter $4\times10^{-4}$ to train ResNet-32 on CIFAR-10. The \texttt{total\_epochs} is set to $200$.
    \item We use the optimizer with batch size $256$, \texttt{base\_lr} $0.2$ and weight decay parameter $5\times10^{-4}$ to train WideResNet-28x10 on CIFAR-100. The \texttt{total\_epochs} is set to $300$.
    \item We use the optimizer with batch size $512$, \texttt{base\_lr} $0.4$ and weieght decay parameter $5\times10^{-4}$ to train WideResNet-28x10 on TinyImageNet. The \texttt{total\_epochs} is set to $300$.
\end{itemize}
To obtain \gls{de}-$M$ teacher models, we repeat the described training procedure $M$ times with different random seeds.
For \gls{be}-$M$ models, we follow the latest official implementation\footnote{\href{https://github.com/google/uncertainty-baselines/tree/ffa818a665655c37e921b411512191ad260cfb47}{https://github.com/google/uncertainty-baselines/tree/ffa818a665655c37e921b411512191ad260cfb47}}.
In specific, we train all the subnetworks with the same mini-batches; it can be done by repeating the training of a single batch $M$ times during training.
As a result,
\begin{itemize}
    \item We use the optimizer with batch size $128\times4$, \texttt{base\_lr} $0.1$ and weight decay parameter $4\times10^{-4}$ to train \gls{be}-4 for ResNet-32 on CIFAR-10. The \texttt{total\_epochs} is set to $250$, as suggested in \citet{wen2019batchensemble}.
    \item We use the optimizer with batch size $64\times8$, \texttt{base\_lr} $0.05$ and weight decay parameter $4\times10^{-4}$ to train \gls{be}-8 for ResNet-32 on CIFAR-10. The \texttt{total\_epochs} is set to $250$.
    \item We use the optimizer with batch size $64\times4$, \texttt{base\_lr} $0.05$ and weight decay parameter $5\times10^{-4}$ to train \gls{be}-4 for WideResNet models on CIFAR-100 and TinyImageNet.
\end{itemize}

\paragraph{Hyperparameters for knowledge distillation}
We searched for several $(\alpha,\tau)$ pairs via grid search to run our experiments with the best hyperparameters.
As shown in \cref{tab:tune_hparams_r32}, we did not find a significant effect of different hyperparameter settings on performance for ResNet-32 on CIFAR-10. 
We therefore decided to stay consistent with the convention, \ie, $\alpha=0.9$ and $\tau=4$.
However, as shown in \cref{tab:tune_hparams_wrn28x10}, we empirically found that $\tau=4$ is not suitable to perform knowledge distillation for WideResNet-28x10 on CIFAR-100, and decided to use $\tau=1$ which achieved the best \gls{nll} across temperature values.
From this, we decided to use $\alpha=0.9$ and $\tau=4$ for WideResNet models on CIFAR-100 and TinyImageNet.

\begin{table}[t]
    \footnotesize
    \caption{Validation \gls{acc} and calibrated \gls{nll} for different values of hyperparameters $\alpha$ and $\tau$ for knowledge distillation from \gls{de}-4 into \gls{be}-4 of ResNet-32 on CIFAR-10.
    All values are measured four times.\\}
    \label{tab:tune_hparams_r32}
    \centering
    \begin{tabular}{lllllll}
    \toprule
    & \multicolumn{3}{c}{\gls{acc}@Valid} & \multicolumn{3}{c}{\gls{nll}@Valid} \\
    \cmidrule(lr){2-4}\cmidrule(lr){5-7}
    & $\alpha=0.8$ & $\alpha=0.9$ & $\alpha=1.0$ & $\alpha=0.8$ & $\alpha=0.9$ & $\alpha=1.0$ \\
    \midrule
    $\tau=2$ & $94.67\spm{0.18}$ & $94.68\spm{0.17}$ & $94.59\spm{0.16}$ & $0.169\spm{0.005}$ & $0.170\spm{0.003}$ & $0.171\spm{0.003}$ \\
    $\tau=3$ & $94.64\spm{0.16}$ & $94.59\spm{0.11}$ & $94.57\spm{0.11}$ & $0.172\spm{0.003}$ & $0.173\spm{0.003}$ & $0.174\spm{0.006}$ \\
    $\tau=4$ & $94.48\spm{0.20}$ & $94.60\spm{0.14}$ & $94.58\spm{0.15}$ & $0.172\spm{0.002}$ & $0.174\spm{0.003}$ & $0.176\spm{0.001}$ \\
    $\tau=5$ & $94.50\spm{0.22}$ & $94.46\spm{0.15}$ & $94.45\spm{0.20}$ & $0.172\spm{0.001}$ & $0.175\spm{0.004}$ & $0.176\spm{0.005}$ \\
    \bottomrule
    \end{tabular}
    \vspace{1em}
    \footnotesize
    \caption{Validation \gls{acc} and calibrated \gls{nll} for different values of hyperparameters $\tau$ for knowledge distillation from \gls{de}-4 into \gls{be}-4 of WideResNet-28x10 on CIFAR-100 where $\alpha$ is fixed to $0.9$.
    All values are measured one time.\\}
    \label{tab:tune_hparams_wrn28x10}
    \centering
    \begin{tabular}{rlllll}
    \toprule
                 & $\tau=1$ & $\tau=2$ & $\tau=3$ & $\tau=4$ & $\tau=5$ \\
    \midrule
    \gls{acc}@Valid & $80.29$ & $80.32$ & $80.08$ & $80.52$ & $80.98$ \\
    \gls{nll}@Valid & $0.757$ & $0.835$ & $0.831$ & $0.827$ & $0.818$ \\
    \bottomrule
    \end{tabular}
\end{table}

\subsection{Evaluation}
\label{subsec:app_evaluation}

For further metric descriptions, we denote a neural network as $\calF(\bsx) : \bbR^D \rightarrow [0, 1]^K$; $\calF$ outputs class probabilities over $K$ classes, and we denote the logits before softmax as $\hat{\calF}(\bsx)$.

\paragraph{Standard metrics}
For a model $\calF$, standard metrics including accuracy, negative log-likelihood, Brier score and expected calibration error are defined as follows:
\begin{itemize}
    \item Accuracy (\gls{acc}):
    \[
    \text{ACC}(\calF)
    = \Pr_{(\bsx, \bsy) \in \calD} \left[
        \argmax_{k \in \{1...K\}}{\left\{\calF^{(k)}(\bsx)\right\}} = \argmax_{k \in \{1...K\}}{\left\{\bsy^{(k)}\right\}}
    \right].
    \]
    \item Negative Log-Likelihood (\gls{nll}):
    \[
    \text{NLL}(\calF)
    = \bbE_{(\bsx, \bsy) \in \calD} \left[
        - \sum_{k=1}^{K} \bsy^{(k)} \log{\calF^{(k)}(\bsx)}
    \right].
    \]
    \item Brier Score (\gls{bs}):
    \[
    \text{BS}(\calF)
    = \bbE_{(\bsx, \bsy) \in \calD} \left[
        \frac{1}{K} \sum_{k=1}^{K} \left(\calF^{(k)}(\bsx) - \bsy^{(k)}\right)^2
    \right].
    \]
    \item Expected Calibration Error (\gls{ece}):
    \[
    \text{ECE}(\calF)
    = \sum_{l=1}^{L} \frac{|B_l|}{N} \bigg|
        \text{acc}(B_l) - \text{conf}(B_l)
    \bigg|,
    \]
    where $N$ is the total number of examples; $|B_l|$ denotes the number of predictions in $l$th bin; $\text{acc}(B_l)$ and $\text{conf}(B_l)$ respectively denote the mean accuracy and mean confidence of predictions in $l$th bin.
\end{itemize}

\paragraph{Calibrated metrics}
For a model $\calF$, we first compute the optimal temperature which minimizes a negative log-likelihood over validation set $\calD_{\textrm{val}}$ as
\[
\tau^\ast
= \argmin_{\tau > 0}{\left\{
        \bbE_{(\bsx, \bsy) \in \calD_{\textrm{val}}} \left[
            - \sum_{k=1}^{K} \bsy^{(k)} \log{ \softmax{ \left( \hat{\calF}^{(k)}(\bsx) / \tau  \right) } }
        \right]
    \right\}},
\]
and then, compute evaluation metrics with temperature scaled outputs:
\[
\calF^{(k)}(\bsx \; ; \; \tau^\ast) \gets \softmax{\left( \hat{\calF}^{(k)}(\bsx) / \tau^\ast \right)},
\quad \text{where } k \in \{1,\dots,K\}.
\]

\paragraph{Evaluation for ensemble models}
\gls{de} and \gls{be} construct a prediction by averaging the outputs of ensemble members (subnetworks) as follows:
\[
\calF^{(k)}(\bsx) \gets \sum_{m=1}^{M} \calF_{m}^{(k)}(\bsx) / M,
\]
where $k \in \{1...K\}$ and $\calF_m$ denotes $m$th ensemble member.
Therefore, we define the logits of the ensemble predictions as
\[
\hat{\calF}^{(k)}(\bsx) \gets \log{\left( \sum_{m=1}^{M} \calF_{m}^{(k)}(\bsx) / M \right)}, \quad \text{where } k \in \{1...K\},
\]
with which both standard and calibrated metrics can be computed.
Also, we also compute the Deep Ensemble Equivalent (\gls{dee}) score~\citep{ashukha2020pitfalls}; for a model $\calS$, \gls{dee} score is defined as
\[
    \text{DEE}(\calS) = \min\{
        \ell \geq 0 \,|\, \text{NLL}(\text{Ensemble of } \ell \text{ models}) \leq \text{NLL}(\calS)
    \}.
\]
That is, \gls{dee} computes the minimum number of independent ensemble members needed to achieve the same performance as a given model $\calS$.
The \gls{nll} values of ensembles with non-integer $\ell$ values are obtained by linear interpolation.
Refer to \cref{tab:table_kd_std} and \cref{tab:table_kd_cross_std} for evaluation results with standard metrics, and \cref{tab:table_kd} and \cref{tab:table_kd_cross} for evaluation results with calibrated metrics.

\begin{table}[t]
    \caption{Knowledge distillation from \gls{de}-$M$ into \gls{be}-$M$ where $M$ denotes the size of ensembles:
    \gls{acc} and \textit{standard metrics} including \gls{nll}, \gls{bs}, \gls{ece}, and \gls{dee}.
    Refer to \cref{tab:table_kd} for the results with calibrated metrics.\\}\label{tab:table_kd_std}
    \centering
    \resizebox{0.95\linewidth}{!}{ 
    \begin{tabular}{lrlllll}
    \toprule
    & & \multicolumn{5}{c}{\textbf{BatchEns << DeepEns (ResNet-32 on CIFAR-10)}} \\
    \cmidrule(r){3-7}
    Method                            &  \# Params & ACC               & NLL                & BS                 & ECE                & DEE                \\
    \midrule
    $\calT$ : DeepEns-4               &   $1.86$ M & $94.42$           & $0.170$            & $0.081$            & $0.008$            & -                  \\
    $\calS$ : BatchEns-4              &   $0.47$ M & $93.37\spm{0.11}$ & $0.282\spm{0.003}$ & $0.107\spm{0.001}$ & $0.039\spm{0.001}$ & $0.994\spm{0.001}$ \\
    \hspace{4mm}+ KD                  &            & $93.98\spm{0.20}$ & $0.252\spm{0.005}$ & $0.099\spm{0.002}$ & $0.038\spm{0.001}$ & $1.261\spm{0.074}$ \\
    \hspace{4mm}+ KD + Gaussian       &            & $93.93\spm{0.12}$ & $0.248\spm{0.004}$ & $0.099\spm{0.002}$ & $0.038\spm{0.002}$ & $1.316\spm{0.062}$ \\
    \tikzmark{startreccc}\hspace{4mm}+ KD + ODS            &            & $93.89\spm{0.10}$ & $\mathbf{0.206}\spm{0.004}$ & $0.094\spm{0.001}$ & $\mathbf{0.028}\spm{0.001}$ & $\mathbf{1.938}\spm{0.067}$ \\
    \hspace{4mm}+ KD + ConfODS        &            & $\mathbf{94.01}\spm{0.19}$ & $0.211\spm{0.002}$ & $\mathbf{0.093}\spm{0.001}$ & $0.029\spm{0.001}$ & $1.865\spm{0.036}$ \tikzmark{endreccc}\\
    \midrule
    $\calT$ : DeepEns-8               &   $3.71$ M & $94.78$           & $0.157$            & $0.077$            & $0.004$            & -                  \\
    $\calS$ : BatchEns-8              &   $0.48$ M & $93.47\spm{0.14}$ & $0.263\spm{0.011}$ & $0.104\spm{0.003}$ & $0.035\spm{0.002}$ & $1.123\spm{0.123}$ \\
    \hspace{4mm}+ KD                  &            & $94.15\spm{0.13}$ & $0.243\spm{0.003}$ & $0.095\spm{0.001}$ & $0.035\spm{0.002}$ & $1.394\spm{0.039}$ \\
    \hspace{4mm}+ KD + Gaussian       &            & $94.09\spm{0.08}$ & $0.245\spm{0.003}$ & $0.097\spm{0.001}$ & $0.037\spm{0.000}$ & $1.361\spm{0.045}$ \\
    \tikzmark{startrecbb}\hspace{4mm}+ KD + ODS            &            & $94.13\spm{0.08}$ & $\mathbf{0.200}\spm{0.005}$ & $\mathbf{0.090}\spm{0.001}$ & $\mathbf{0.028}\spm{0.001}$ & $\mathbf{2.105}\spm{0.158}$ \\
    \hspace{4mm}+ KD + ConfODS        &            & $\mathbf{94.18}\spm{0.12}$ & $0.205\spm{0.004}$ & $0.091\spm{0.001}$ & $0.029\spm{0.001}$ & $1.950\spm{0.066}$ \tikzmark{endrecbb}\\
    \midrule
    \midrule
    & & \multicolumn{5}{c}{\textbf{BatchEns << DeepEns (WideResNet-28x10 on CIFAR-100)}} \\
    \cmidrule(r){3-7}
    Method                            &  \# Params & ACC               & NLL                & BS                 & ECE                & DEE                \\
    \midrule
    $\calT$ : DeepEns-4               & $146.15$ M & $82.52$           & $0.676$            & $0.250$            & $0.035$            & -                  \\
    $\calS$ : BatchEns-4              &  $36.62$ M & $80.34\spm{0.08}$ & $0.900\spm{0.005}$ & $0.298\spm{0.002}$ & $0.094\spm{0.001}$ & $0.972\spm{0.001}$ \\
    \hspace{4mm}+ KD                  &            & $80.51\spm{0.22}$ & $0.803\spm{0.010}$ & $0.283\spm{0.001}$ & $0.073\spm{0.005}$ & $1.000\spm{0.007}$ \\
    \hspace{4mm}+ KD + Gaussian       &            & $80.39\spm{0.12}$ & $0.816\spm{0.010}$ & $0.286\spm{0.000}$ & $0.070\spm{0.003}$ & $0.994\spm{0.003}$ \\
    \tikzmark{startrecaa}\hspace{4mm}+ KD + ODS            &            & $\mathbf{81.88}\spm{0.32}$ & $0.680\spm{0.016}$ & $\mathbf{0.257}\spm{0.006}$ & $\mathbf{0.023}\spm{0.003}$ & $3.987\spm{1.325}$ \\
    \hspace{4mm}+ KD + ConfODS        &            & $81.85\spm{0.32}$ & $\mathbf{0.678}\spm{0.009}$ & $\mathbf{0.257}\spm{0.003}$ & $\mathbf{0.023}\spm{0.001}$ & $\mathbf{3.990}\spm{0.770}$ \tikzmark{endrecaa}\\
    \bottomrule
    \begin{tikzpicture}[remember picture,overlay]
    \foreach \Val in {recaa,recbb,reccc}
    {
    \draw[rounded corners,black,dotted]
      ([shift={(1.5ex,2ex)}]pic cs:start\Val) 
        rectangle 
      ([shift={(0.5ex,-0.5ex)}]pic cs:end\Val);
    }
    \end{tikzpicture}
    \end{tabular} }
\end{table}

\begin{table}[t]
    \footnotesize
    \setlength{\tabcolsep}{1em}
    \caption{Cross-architecture knowledge distillation for a model compression on CIFAR-100 and TinyImageNet: \gls{acc} and \textit{standard metrics} including \gls{nll}, \gls{bs}, \gls{ece}, and \gls{dee}. Refer to \cref{tab:table_kd_cross} for the results with calibrated metrics.\\}
    \label{tab:table_kd_cross_std}
    \centering
    \resizebox{0.95\linewidth}{!}{
    \begin{tabular}{lrllll}
    \toprule
    & & \multicolumn{4}{c}{\textbf{BatchEns-4 << DeepEns-4 (WRN-28x2 on CIFAR-100)}} \\
    \cmidrule(r){3-6}
    Method                         &  \# Params & \gls{acc} ($\uparrow$) & \gls{nll} ($\downarrow$) & \gls{bs} ($\downarrow$) & \gls{ece} ($\downarrow$) \\
    \midrule
    $\calT$ : DeepEns-4            & $146.15$ M & $82.52$           & $0.676$            & $0.250$            & $0.035$ \\
    $\calS$ : BatchEns-4           &   $1.50$ M & $75.17\spm{0.27}$ & $1.245\spm{0.024}$ & $0.383\spm{0.004}$ & $0.141\spm{0.004}$ \\
    \hspace{4mm}+ KD               &            & $75.19\spm{0.36}$ & $1.207\spm{0.021}$ & $0.377\spm{0.007}$ & $0.136\spm{0.005}$ \\
    \hspace{4mm}+ KD + Gaussian    &            & $74.50\spm{0.17}$ & $1.247\spm{0.012}$ & $0.389\spm{0.004}$ & $0.137\spm{0.003}$ \\
    \tikzmark{startrecff}\hspace{4mm}+ KD + ODS         &            & $\mathbf{76.03}\spm{0.22}$ & $\mathbf{0.899}\spm{0.013}$ & $\mathbf{0.333}\spm{0.003}$ & $\mathbf{0.027}\spm{0.002}$ \\
    \hspace{4mm}+ KD + ConfODS     &            & $76.01\spm{0.16}$ & $0.901\spm{0.006}$ & $0.334\spm{0.002}$ & $0.028\spm{0.004}$ \tikzmark{endrecff}\\
    \midrule
    & & \multicolumn{4}{c}{\textbf{BatchEns-4 << DeepEns-4 (WRN-28x5 on CIFAR-100)}} \\
    \cmidrule(r){3-6}
    Method                         &  \# Params & \gls{acc} ($\uparrow$) & \gls{nll} ($\downarrow$) & \gls{bs} ($\downarrow$) & \gls{ece} ($\downarrow$) \\
    \midrule
    $\calT$ : DeepEns-4            & $146.15$ M & $82.52$           & $0.676$            & $0.250$            & $0.035$ \\
    $\calS$ : BatchEns-4           &   $9.20$ M & $78.75\spm{0.11}$ & $1.031\spm{0.022}$ & $0.324\spm{0.004}$ & $0.112\spm{0.003}$ \\
    \hspace{4mm}+ KD               &            & $78.89\spm{0.10}$ & $0.932\spm{0.021}$ & $0.314\spm{0.004}$ & $0.095\spm{0.001}$ \\
    \hspace{4mm}+ KD + Gaussian    &            & $78.80\spm{0.41}$ & $0.929\spm{0.018}$ & $0.313\spm{0.007}$ & $0.090\spm{0.006}$ \\
    \tikzmark{startrecee}\hspace{4mm}+ KD + ODS         &            & $80.24\spm{0.05}$ & $0.744\spm{0.007}$ & $0.279\spm{0.002}$ & $0.026\spm{0.003}$ \\
    \hspace{4mm}+ KD + ConfODS     &            & $\mathbf{80.62}\spm{0.25}$ & $\mathbf{0.735}\spm{0.007}$ & $\mathbf{0.275}\spm{0.003}$ & $\mathbf{0.023}\spm{0.001}$ \tikzmark{endrecee}\\
    \midrule
    \midrule
    & & \multicolumn{4}{c}{\textbf{BatchEns-4 << DeepEns-4 (WRN-28x5 on TinyImageNet)}} \\
    \cmidrule(r){3-6}
    Method                         &  \# Params & \gls{acc} ($\uparrow$) & \gls{nll} ($\downarrow$) & \gls{bs} ($\downarrow$) & \gls{ece} ($\downarrow$) \\
    \midrule
    $\calT$ : DeepEns-4            & $146.40$ M & $69.90$ & $1.243$ & $0.404$ & $0.027$ \\
    $\calS$ : BatchEns-4           &   $9.23$ M & $64.86$ & $2.019$ & $0.520$ & $0.179$ \\
    \hspace{4mm}+ KD               &            & $65.86$ & $1.782$ & $0.493$ & $0.149$ \\
    \hspace{4mm}+ KD + Gaussian    &            & $65.72$ & $1.792$ & $0.494$ & $0.148$ \\
    \tikzmark{startrecdd}\hspace{4mm}+ KD + ODS         &            & $\mathbf{65.98}$ & $\mathbf{1.440}$ & $\mathbf{0.460}$ & $\mathbf{0.059}\color{white}\spm{0.000}$ \tikzmark{endrecdd}\\
    \bottomrule
    \begin{tikzpicture}[remember picture,overlay]
    \foreach \Val in {recdd,recee,recff}
    {
    \draw[rounded corners,black,dotted]
      ([shift={(1.5ex,2ex)}]pic cs:start\Val) 
        rectangle 
      ([shift={(0.5ex,-0.5ex)}]pic cs:end\Val);
    }
    \end{tikzpicture}
    \end{tabular}}
\end{table}


\end{document}